\documentclass[journal]{IEEEtran}
\pdfoutput=1
\usepackage{float}
\usepackage{graphicx}
\usepackage{subcaption}
\usepackage{textcomp}
\usepackage{xcolor}
\usepackage{float}
\usepackage{multirow}
\usepackage{csquotes}
\usepackage{url}
\usepackage{amsmath}
\usepackage{booktabs}

\usepackage{tabularx}
\usepackage{booktabs}
\newcolumntype{Y}{>{\raggedleft\let\newline\\\arraybackslash\hspace{0pt}}X}
\newcolumntype{Z}{>{\centering\let\newline\\\arraybackslash\hspace{0pt}}X}

\usepackage{enumitem}
\setlist[itemize]{noitemsep,topsep=0pt}

\ifCLASSINFOpdf

\else

\fi

\begin{document}

\title{Selfie Periocular Verification using an Efficient Super-Resolution Approach}

\author{Juan Tapia,~\IEEEmembership{Member,~IEEE,}
        Andres Valenzuela,~\IEEEmembership{Member,~IEEE,}
        Rodrigo Lara,~\IEEEmembership{Member,~IEEE,}\\
        Marta Gomez-Barrero,~\IEEEmembership{Member,~IEEE,}
        and~Christoph Busch,~\IEEEmembership{Senior Member,~IEEE,}\\ 
\thanks{Juan Tapia and Christoph Busch are with the da/sec-Biometrics and Internet Security Research Group, Hochschule Darmstadt, Germany, e-mail: \{juan.tapia-farias, christoph.busch\}@h-da.de.}
\thanks{Andres Valenzuela and Rodrigo Lara, R\&D Center are with TOC Biometrics Company, Santiago, Chile, email: \{andres.valenzuela, rodrigo.lara\}@tocbiometrics.com}%
\thanks{Marta Gomez-Barrero is with the Hochschule Ansbach, Germany, email: marta.gomez-barrero@hs-ansbach.de}
\textbf{**The following paper is a pre-print- The publication is currently under revision process}
\thanks{Manuscript received xxx; revised xx.}}

\markboth{Journal of \LaTeX\ Class Files,~Vol.~14, No.~8, August~2015}%
{Shell \MakeLowercase{\textit{et al.}}: Bare Demo of IEEEtran.cls for IEEE Journals}

\maketitle

\begin{abstract}
Selfie-based biometrics has great potential for a wide range of applications since, e.g. periocular verification is contactless and is safe to use in pandemics such as COVID-19, when a major portion of a face is covered by a facial mask. Despite its advantages, selfie-based biometrics presents challenges since there is limited control over data acquisition at different distances. Therefore, Super-Resolution (SR) has to be used to increase the quality of the eye images and to keep or improve the recognition performance. 
We propose an Efficient Single Image Super-Resolution algorithm, which takes into account a trade-off between the efficiency and the size of its filters. To that end, the method implements a loss function based on the Sharpness metric used to evaluate iris images quality. Our method drastically reduces the number of parameters compared to the state-of-the-art: from 2,170,142 to 28,654. Our best results on remote verification systems with no redimensioning reached an EER of 8.89\% for FaceNet, 12.14\% for VGGFace, and 12.81\% for ArcFace. Then, embedding vectors were extracted from SR images, the FaceNet-based system yielded an EER of 8.92\% for a resizing of x2, 8.85\% for x3, and 9.32\% for x4.
\end{abstract}


\IEEEpeerreviewmaketitle

\section{Introduction}
\label{sec:intro}

\IEEEPARstart{S}martphones, and mobile devices in general, play nowadays a central role in our society. We use them on a daily basis not only for communication purposes, but also to access social media and for sensitive tasks such as online banking. In order to increase the security level of those more sensitive applications, verifying the subject's identity plays a key role. To tackle this requirement, many companies are currently working towards creating applications to verify the subject's identity by comparing a selfie image with the reference face image stored in the embedded chip of an ID-Card/Passport and a selfie image using Near Field Communication (NFC) from smartphones \cite{fake-id}. This represents a user-friendly identity verification process, which can be easily embedded into numerous applications. However, this verification process also faces some challenges: for instance, that selfie image is captured in an uncontrolled scenario, where occlusions due to wearing a scarf in winter or a hygienic facial mask during a pandemic such as COVID-19 may hinder the performance of general face recognition algorithms. Therefore, there is a reinforced need to explore alternatives which can deal with those occluded images successfully, such as utilising the periocular region for recognition purposes.

The aforementioned reasons have increased the interest on periocular based biometrics in the last decade in different scenarios ~\cite{Alonso-SurveyPeriocular-PRL-2016,alonso2019cross, naser, Raja2014BinarizedSF}. In particular, it has been shown that periocular images captured with mobile devices for recognition purposes are mainly acquired as selfie face images. And the number of digital photos will increase every year: in 2022, 1.5 trillion images were taken, and 90\% of them come from smartphones\footnote{\url{https://blog.mylio.com/how-many-photos-taken-in-2022/}}. In order to recognise individuals from a selfie in a remote verification system, the periocular region needs to be cropped, and the resulting periocular sample has often a very low-resolution \cite{Tapia2019}. Moreover, the subjects capture selfie images in multiple places and backgrounds, using selfie sticks, alone, or with others. This translates into a high intra-class variability which can be observed for the images, in terms of size, lighting conditions, and face pose.

\begin{figure*}[t]
\centering
\includegraphics[width=0.80\textwidth]{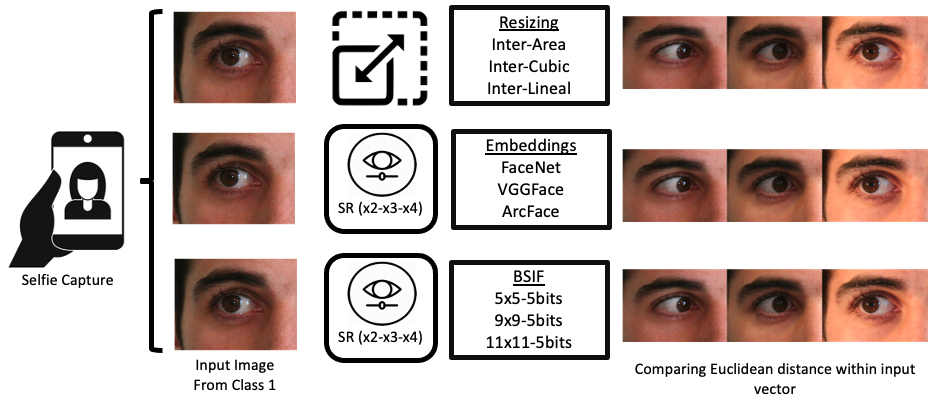}
\caption{\label{workflow} Block diagram of the verification system proposed, including a super-resolution approach. Top: Traditional approach with resizing images. SR approach with deep learning embeddings (Middle) and with handcrafted features (BSIF, Bottom).}
\end{figure*}

With the aim of improving the quality of such low-resolution images, several Single Image Super-Resolution (SISR) methods have been recently proposed ~\cite{wang_survey, KUMARI2019, Yang, Tian}, mainly based on convolutional neural networks (CNNs). Even though some authors have enhanced such networks to achieve more efficiently the reconstruction results of the super-resolution \cite{TimofteRG15}, most approaches still use deep models, which demand large resources and are thus not suitable for mobile or Internet-of-Things (IoT) devices. Furthermore, the loss function used in most techniques is based on structural similarity (SSIM) and Peak Signal to Noise Ratio (PSNR) metrics \cite{wang_survey}. Even though those metrics are appropriate for increasing the resolution of general purpose images (e.g., landscapes, cities, or birds) they are not that suitable for increasing the quality of iris based biometrics applications. In contrast, the ISO/IEC 29794 standard on biometric sample quality — Part 6: Iris image data describes sharpness based on the Laplacian of Gaussian (LoG) as one relevant quality. 

In this work, we have a twofold goal: verify a biometric claim in a verification transaction from a smartphone selfie periocular image in the visible spectrum (VIS) and propose an efficient super-resolution approach (see Fig.~\ref{workflow}). As already mentioned, this is a challenging task since there is limited control of the quality of the images taken: selfies can be captured from different distances, light conditions, and resolutions. Therefore, to tackle these issues, we present a SISR algorithm with a novel loss function based on the sharpness LoG metric and a light-weight CNN. This model takes into account the trade-off between the number of layers and filter sizes in order to achieve a light model suitable for mobile devices. Additionally, we explore pixel-shuffle and transposed convolutions in order to recover the fine details of the periocular eye images. To validate our approach, we use different databases for training and testing. In addition, we benchmark both handcrafted features and pre-trained deep learning models. Our method drastically reduces the number of parameters when compared with the state-of-the-art Deep CNNs with Skip Connection and Network (DCSCN)~\cite{yamanaka}: from 2,170,142 to 28,654 parameters when the image size is increased by a factor of 2. 

This paper is an extension of our previous work~\cite{tapia_wifs2020}. In that work, we focused on achieving an accurate Enhanced SISR (ESISR) algorithm for periocular eye images taken from selfie images, reporting results in terms of image similarity for the recovered images on a smaller Samsung dataset. In this paper, we evaluate this new ESISR architecture in more detail and benchmark it with two new state-of-the-art methods: WDSR-A~\cite{yu} and SRGAN~\cite{lim}. A full explanation of the reasons that led us to such architecture is discussed in this work. As an additional contribution, this manuscript includes the performance evaluation of our proposed methods on periocular verification systems using three pre-trained CNNs: FaceNet~\cite{facenet}, VGGFace~\cite{Parkhi15}, and ArcFace~\cite{deng2018arcface}. All methods have been now evaluated on the larger MobBIO~\cite{Sequeira} and NTNU~\cite{visper} databases. A benchmark with a traditional resized method such as inter-area, inter-lineal, and inter-cubic (bicubic) has been also analysed. A handcrafted feature extractor, Binary Statistical Image Filter (BSIF), was also added to evaluate and compare the results with the deep learning approach. Detection Error Trade-off (DET) curves are included to show our proposal's performance and efficiency. All these new experiments are benchmarked with those previously obtained in \cite{yamanaka, yu, ledig}.

Therefore, the main contributions from this work can be summarised as follows:
\begin{itemize}
    \item An efficient SR architecture is proposed, using only seven layers with a feature extractor and one block based on recursive learning of reconstruction.
    
    \item A recursive pixel-shuffle technique is introduced over a transposed convolution to extract and keep fine details of periocular images.
    
    \item  A novel loss function that includes the LoG sharpness iris quality metric and the SR loss function was proposed.
    
    \item A significant reduction of the number of parameters in comparison with the state-of-the-art using WDSR-A, SRGAN and DCSCN algorithms (see Sect.~\ref{sec:relate}) is reported.
   
    \item A novel database for selfie periocular eye images was acquired and is available for researchers upon request.
    
    \item A periocular verification system based on an embedded vector from three pre-trained models (FaceNet, VGGFace, and ArcFace), with an SR-based pre-processing of the samples (x2, x3 and x4) was tested.
    
    \item A benchmark between deep learning approaches and a handcrafted method is reported.
    
    \item A full analysis of the influence of SR on selfie biometrics scenarios with traditional resizing methods (Interlineal, InterCubic, InterArea) was also included.
     
\end{itemize}

The rest of the article is organised as follows. Sect.\ref{sec:relate} summarises the related works on periocular recognition and super resolution. The new recognition and super-resolution method is described in Sect.~\ref{method}. The experimental framework is then presented in Sect.~\ref{experiments} and the results are discussed in Sect.~\ref{sec:results}. We conclude the article in Sect.~\ref{conclusions}.

\section{Related Work}
\label{sec:relate}

\subsection{Super-Resolution (SR)}
Super-resolution (SR) is the process of recovering a high-resolution (HR) image from a low-resolution (LR) one \cite{dong,wang_survey}. Supervised machine learning approaches learn mapping functions from LR images to HR images from a large number of examples. The mapping function learned by these models is the inverse of a downgrade function that transforms HR images into LR images. Such downgrade functions can be known or unknown.

Many state-of-the-art SR models learn most of the mapping function in LR space followed by one or more upsampling layers at the end of the network. This is called post-upsampling. Earlier approaches first upsampled the LR image with a pre-defined up-sampling operation and then learned the mapping in the HR space (pre-upsampling SR). A disadvantage of this approach is that more parameters per layer are required, which in turn leads to higher computational costs and limits the construction of deeper neural networks \cite{wang_survey}. SR requires that most of the information contained in an LR image must be preserved in the SR image. SR models therefore mainly learn the residuals between LR and HR images. Residual network designs are therefore of high importance: identity information is conveyed via skip connections whereas reconstruction of high frequency content is done on the main path of the network \cite{wang_survey}.

Dong \textit{et al.}~\cite{dong} proposed several SISR algorithms which can be categorized into four types: prediction models, edge-based methods, image statistical methods, and patch-based (or example-based) methods. This method uses 2 to 4 convolutional layers to prove that the learned model performs well on SISR tasks. The authors concluded that \emph{using a larger filter size is better than using deeper Convolutional Neural Networks (CNNs)}.

Kim \textit{et al.}~\cite{Kim} proposed an image SR method using a Deeply-Recursive Convolutional Network (DRCN), which contains deep CNNs with up to 20 layers. Consequently, the model has a huge number of parameters. However, the CNNs share each other's weights to reduce the number of parameters to be trained, thereby being able to succeed in training the deep CNN network and achieving a significant performance. The authors conclude in their work \emph{that deeper networks are better than large filters}.

Yamanaka \textit{et al.}~\cite{yamanaka} proposed a Deep CNN with a Residual Net, Skip Connection and Network (DCSCN) model achieving a state- of-the-art reconstruction performance while reducing by at least 10 times the computational cost. According to the existing literature, deep CNNs with residual blocks and skip connections are suitable to capture fine details in the reconstruction process. In the same context, \cite{shi} and \cite{long} propose the pixel-shuffle and transposed convolution algorithm in order to extract the most relevant features from the images. The transposed convolutional layer can learn up-sampling kernels. However, the process is similar to the usual convolutional layer and the reconstruction ability is limited. To obtain a better reconstruction performance, the transposed convolutional layers need to be stacked, which means the whole process needs high computational resources \cite{yamanaka}. Conversely, pixel-shuffle extracts features from the low-resolution images. The authors \cite{yamanaka} argue that batch normalisation loses scale information of images and reduces the range flexibility of activations. Removal of batch normalisation layers not only increases SR performance but also reduces GPU memory 40\%. This way, significantly larger models can be trained.

Ledig \textit{et al.}~\cite{ledig} proposed a deep residual network which is able to recover photo-realistic textures from heavily downsampled images on public benchmarks. An extensive Mean-Opinion-Score (MOS) test shows significant gains in perceptual quality using SR based on Generative Adversarial Network (SRGAN). In addition, the authors present a new perceptual loss based on content loss and adversarial loss.
 
Yu \textit{et al.}~\cite{yu} proposed the key idea of wide activation to explore efficient ways to expand features before ReLU, since simply adding more parameters is inefficient for smartphone based image SR scenarios. The authors present two new networks named Wide Activation for Efficient and Accurate Image Super-Resolution (WSDR). These networks (WDSR-A and WDSR-B) yielded better results on the large-scale DIV2K image super resolution benchmark in terms of PSNR with the same or lower computational complexity. Similar results but with a larger number of parameters are presented by Lim \textit{et al.}~\cite{lim} in a model called Enhanced Deep Residual Networks for Single Image Super Resolution (EDSR).

Specifically for biometric applications, some papers have explored the use of SR in iris recognition in the visible and near-infrared spectrum. Ribeiro \textit{et al.}~\cite{ribeiro} proposed a SISR method using CNNs for iris recognition. In particular, the authors test different state- of-the-art CNN architectures and use different training databases in both the near-infrared and visible spectra. Their results are validated on a database of 1,872 near-infrared iris images and on a smartphone image database. The experiments show that using deeper architectures trained with texture databases that provide a balance between edge preservation and the smoothness of the method can lead to good results in the iris recognition process. Furthermore, the authors used PSNR and SSIM to measure the quality of the reconstruction. More recently, Alonso-Fernandez \textit{et al.}~\cite{survey} presented a comprehensive survey of iris SR approaches. They also described an Eigen-patches reconstruction method based on the principal component analysis and Eigen-transformation of local image patches. The inherent structure of the iris is reproduced by building a patch-position-dependent dictionary. The authors also used PSNR and SSIM to measure the quality of the reconstruction in the NIR spectrum and in the NTNU database in the visible spectrum \cite{raja}.

\subsubsection{Metrics}
Deep learning-based methods for SISR significantly outperform conventional approaches in terms of Peak Signal-to-Noise Ratio (PSNR) and Structural Similarity (SSIM) \cite{dong}. In this section, we review these two metrics. 

SSIM is a subjective metric used for measuring the structural similarity between images from the perspective of the human visual system. It is based on three relatively independent properties, namely: luminance, contrast, and structure. The SSIM metric can be seen as a weighted product of the comparison of luminance, contrast, and structure computed independently. Therefore, SSIM is defined as:
\begin{equation}
  \mathrm{SSIM}(x,y) = \frac{(2\mu_x\mu_y + C_1) + (2 \sigma _{xy} + C_2)} 
    {(\mu_x^2 + \mu_y^2+C_1) (\sigma_x^2 + \sigma_y^2+C_2)}
  \label{eq:SSMI}
\end{equation}
where $\mu$ and $\sigma$ represent the average and variance of x and y, respectively; and $C_1$ and $C_2$ are two variables to stabilise the division with a weak denominator.

PSNR is a common objective metric to measure the reconstruction quality of a lossy transformation. It is inversely proportional to the logarithm of the Mean Squared Error (MSE) between the ground truth image and the generated image:
\begin{equation}
    \mathrm{PSNR} = 10 \log_{10}\left(\frac{\max^{2}}{\mathrm{MSE}}\right)
\end{equation}
where max denotes the maximum pixel value, and MSE the mean of the squared of differences between the pixel values of the reconstructed super-resolution image and the ground truth image (prior to downsampling. Therefore, this metric measures pixel differences and not the quality of the images.

\subsection{Periocular recognition}

Periocular recognition based on traditional feature extraction methods such as intensity, shape, texture, fusion, and off-the-shelf CNN features with pre-trained models has been widely studies. However, to the best of our knowledge, only a few papers have explored the use of SR methods to improve the quality of the RGB images coming from periocular selfie captures. 

Chandrashekhar \textit{et al.~}\cite{Chandrashekhar} proposed a new initialization strategy for the definition of the periocular region-of-interest and the performance degradation factor for periocular biometric and the influence of  Histogram of Oriented Gradient (HOG), Local Binary Pattern (LBP), Scale-Invariant Feature Transform (SIFT), Fusion at the Score Level, Effect of Reference Points of the eyes, Covariates, Occlusion Performance and Pigmentation Level Performance.

Raja \textit{et al.~}\cite{kiran2015} explore multi-modal biometrics as a means for secure authentication. The proposed system employs face, periocular, and iris images, all captured with embedded smartphone cameras. As the face image is captured closely, one can always obtain periocular and iris information with fine details. This work also explores various score level fusion schemes of complementary information from all three modalities. Also, the same authors used in~\cite{visper} used in the periocular region for authentication under unconstrained acquisition in biometrics. They acquired a new database named Visible Spectrum Periocular Image (VISPI), and proposed two new feature extraction techniques to achieve robust and blur invariant biometric verification using periocular images captured by smartphones.  

Ahuja \textit{et al.~}\cite{AHUJA201717} proposed a hybrid convolution-based model for verifying pairs of periocular RGB images. They composed a hybrid model as a combination of an unsupervised and a supervised CNN, and augment the combination with SIFT model.

Diaz \textit{et al.~}\cite{PR-Diaz} proposed a method to apply existing architectures pre-trained on the ImageNet Large Scale Visual Recognition Challenge, to the task of periocular recognition. These networks have proven to be very successful for many other computer vision tasks apart from the detection and classification tasks for which they were designed. They demonstrate that these off-the-shelf CNN features can effectively recognise individuals based on periocular images.

More recently, Kumari \textit{et al.~}\cite{KUMARI2019} surveyed periocular biometrics and and provided a deep insight of various aspects, including the periocular region utility as a stand-alone modality, its fusion with iris, its application in the smartphone authentication, and its role in soft biometric classification. In their review, the authors did not mention SR approaches.

\section{Proposed method}
\label{method}
As mentioned in Sect.~\ref{sec:intro} and depicted in Fig.~\ref{workflow}, we focus in this work on a two-stage system. First, we improve the SR approaches for periocular images. Second, we use that improved SR method to enhance the recognition performance of periocular-based biometric systems, in contrast to traditional SR methods. We describe in Sect.~\ref{sec:method:SR} the proposed ESISR technique, and in Sect.~\ref{sec:method:recog} the feature extraction and comparison methods utilised for periocular recognition.

\begin{figure*}[]
\centering
\includegraphics[width=1.0\textwidth]{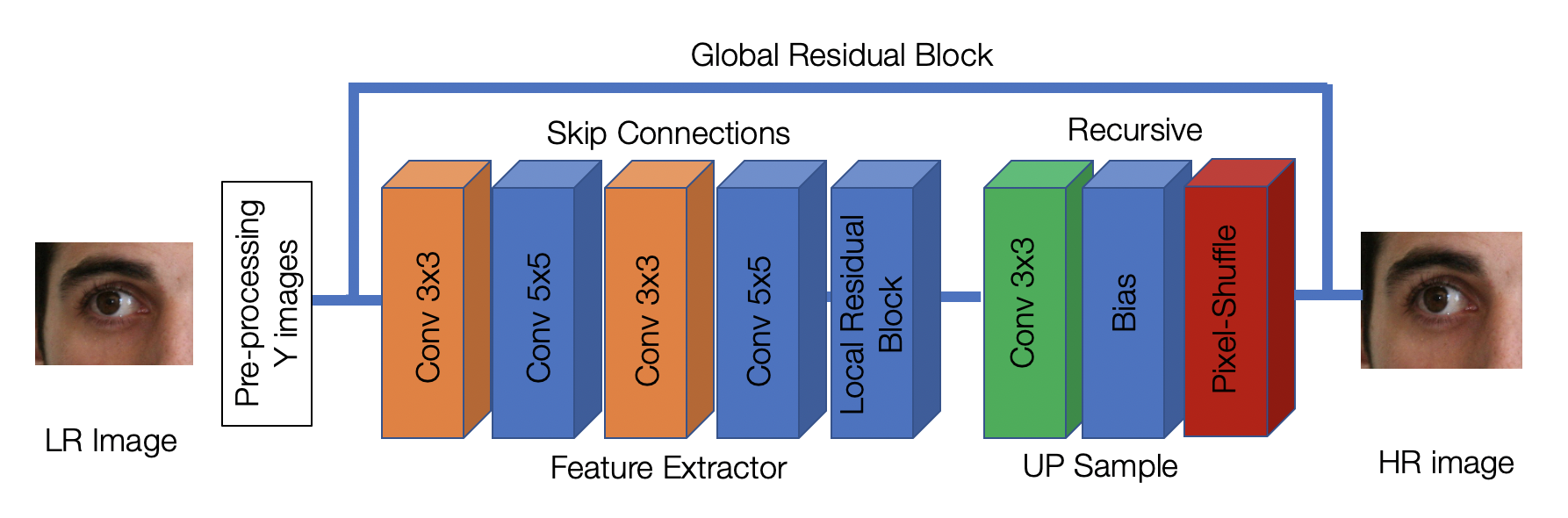}
\caption{\label{block} Proposed ESISR method.}
\end{figure*}

\subsection{Stage-1: Super-Resolution}
\label{sec:method:SR}

In this section, we present an efficient image SR network that is able to recover periocular images from selfies (ESISR). Our network includes two building blocks, as it can be observed in Fig.~\ref{block}: A feature extraction and a reconstruction stage based on DCSCN, which are described in the remaineder of this section. 

Since SR in general is an image-to-image translation task where the input image is highly correlated with the target image, researchers try to learn only the residuals between them (i.e. global residual learning). This process avoids learning a complicated transformation from a complete image to another. Instead, it only requires learning a residual map to restore the missing high-frequency details. Since most regions' residuals are close to zero, the model complexity and learning difficulty are thus greatly reduced. 

This local residual learning is similar to ResNet to alleviate the degradation problem caused by ever-increasing network depths, reduce training difficulty, and improve the learning ability. For these reasons, we are using recursive learning to learn higher-level features without introducing an overwhelming number of parameters, which means applying the same modules multiple times.

In addition to choosing an appropriate network architecture, the definition of the perceptual loss function is critical for the performance of the proposed method based on the DCSCN network, as mentioned in Sects.~\ref{sec:intro} and~\ref{sec:relate}. While SR is commonly based on the MSE, PSNR, and SSMI metrics, we have designed a loss function that incorporates as well a sharpness measure with respect to perceptually relevant features. The function thus balances between reconstructing images by minimising the difference of the sharpness values and weights the results of SSIM and PSNR.

\subsubsection{Pre-processing}

The original RGB images captured with a smartphone represent an additive color-space where colors are obtained by a linear combination of Red, Green, and Blue values. The three channels are thus correlated by the amount of light on the surface. In order to avoid such correlations, all the images were converted from RGB to YCbCr. The YCrCb color space is derived from RGB, and separates the luminance and chrominance components into different channels. In particular, it has the following three components: i) Y, Luminance or Luma component obtained from RGB after gamma correction; ii) $Cr = R – Y$, how far is the red component from Luma; and iii) $Cb = B – Y$, how far is the blue component from Luma. We only use $Y$ component in this work because stored the high resolution luminance information. Instead of CbCr that comprises the image information. The periocular image areas were automatically cropped from faces to the size of $250\times200$ pixels.

\subsubsection{Feature extraction}

\begin{figure}[t]
\centering
\includegraphics[width=0.99\linewidth]{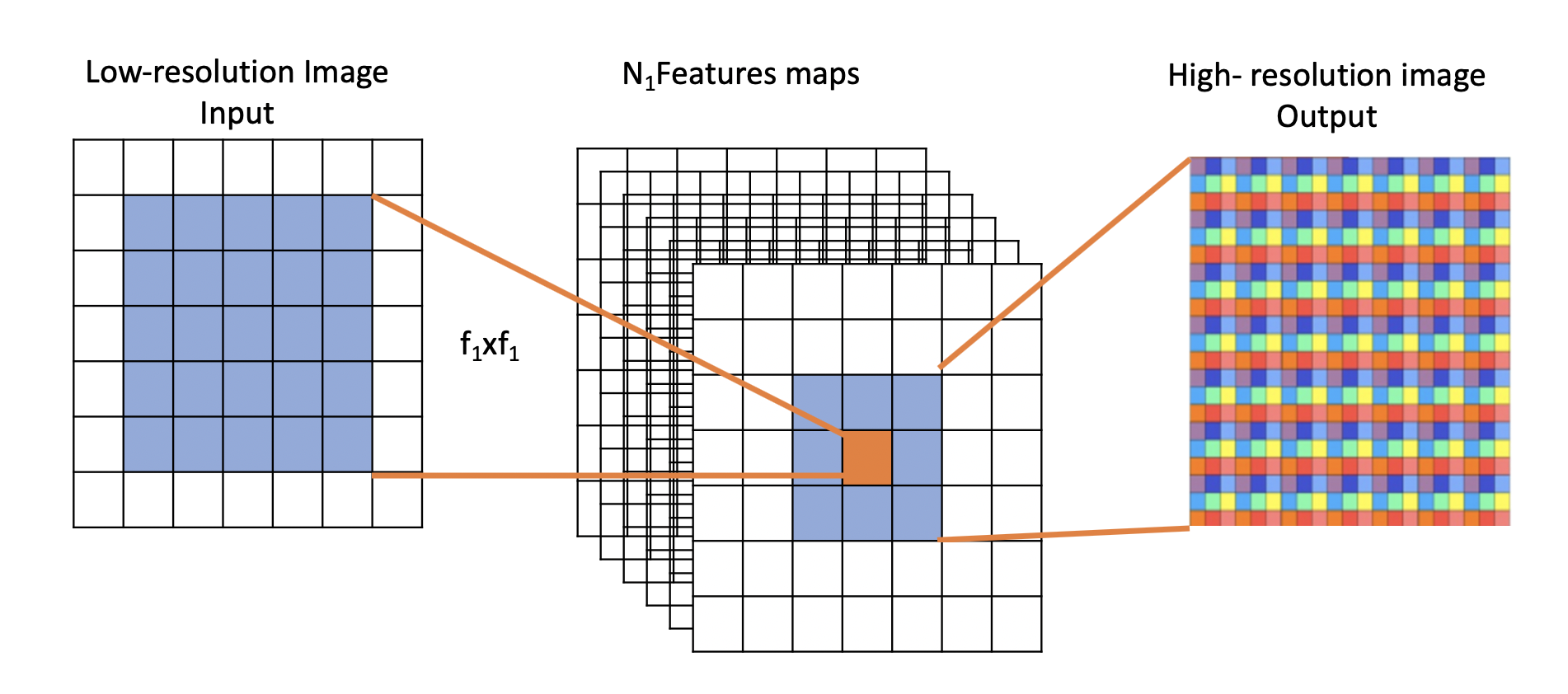}
\caption{\label{pixel} Pixel-shuffle convolution layer that aggregates the feature maps from LR space and builds the SR image in a single step. Based on \cite{ShiCHTABRW16}.}
\end{figure}

As mentioned above, the Y component of the converted image is used as input for our model. Several patches of $32\times32$ and $48\times48$ pixels were extracted from the image and used to grasp the features efficiently. We look for the features that achieve a better trade-off between the number and size of filters of each CNN layer. Seven blocks of $5\times5$ and $3\times3$ have been selected after several experiments. The information is extracted using small convolutional blocks with residual connections and stride convolutions in order to preserve both the global and the fine details in periocular images. Only the final features from $3\times3$ and $5\times5$ pixels are concatenated, following the recursive pixel-shuffle approach (see Fig.~\ref{pixel}). These local skip connections in residual blocks make the network easier to optimise, thereby supporting the construction of deeper networks.

\begin{figure}[t]
\centering
\includegraphics[width=0.99\linewidth]{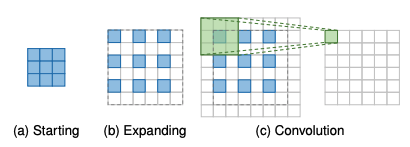}
\caption{\label{transpose} Transpose-convolution operation representation. (a) The starting matrix represents the input image. (b) Expanding operation adds zeros to the images in order to increase the size. c) The convolution operation is performed again in a new resolution. Based on \cite{wang_survey}.}
\end{figure}

A model with transpose convolution instead of pixel-shuffle was trained to explore the quality of the reconstruction images \cite{wang_survey}. See Fig. ~\ref{transpose}. Transpose convolution operates conversely to normal convolution, predicting the input based on feature maps sized like convolution output. It increases image resolution by expanding the image by adding zeros and performing convolution operations.

\subsubsection{Reconstruction}

Our reconstruction stage uses only one convolutional block with 2 layers (Conv + Relu + Conv) in a recursive path. This block includes $3\times3$ convolutions and pixel-shuffle algorithm (see Fig.~\ref{pixel}) to create a high-resolution image from a low-resolution input. Batch normalisation was removed. An optimised sub-pixel convolution layer that learns a matrix of up-scaling filters to increase the final LR feature maps into the SR output was used.

\subsubsection{Perceptual loss function}

The ISO/IEC 29794-6\footnote{\url{https://www.iso.org/standard/54066.html}} on iris image quality introduced a set of quality metrics, that can measure the utility of a sample. Based on the NIST IREX evaluation (footnote:https://www.nist.gov/programs-projects/iris-exchange-irex-overview) a sharpness metric was identified as strongly predictive for recognition performance. We follow this finding and measure:

\begin{equation}
    LoG(x,y)=-\frac{1}{\pi\sigma^4} \left[1- \frac{x^2+y^2}{2\sigma^2}\right]e^{-\frac{x^2+y^2}{2\sigma^2}} \label{eq:sharp}
\end{equation}
The Laplacian of Gaussian operator~(LoG) is thus the sharpness metric used in this work. Calculation of the sharpness of an image is determined by the power resulting from filtering the image with a Laplacian of Gaussian kernel. The standard deviation of the Gaussian is 1,4. 


Now, it is important to highlight that the loss function aims to improve the quality of the reconstruction. To that end, we combine the SSIM and PSNR classical SR metrics with the sharpness metric for iris images recommended, as follows:

\begin{equation}
\begin{split}
L(I_{LR},I_{HR})&= 0.5\cdot \mathrm{LoG}\left(I_{LR}, I_{HR}\right) \cdot 
[0.25\cdot \mathrm{SSIM}\left(I_{LR},I_{HR}\right) 
\\
&+ 0.25\cdot \mathrm{PSNR}\left(I_{LR},I_{HR}\right)]
\label{eq:loss}
\end{split}
\end{equation}

where $I_{LR}$ represents a low-resolution image, $I_{HR}$ the corresponding high-resolution image recovered, and LoG the sharpness as defined in Eq.~\ref{eq:sharp}. The best values of the weights for each specific metric (i.e., 0.25, 0.25 and 0.50) were estimated in a grid search with a train dataset.

\subsection{Stage-2: Periocular recognition}
\label{sec:method:recog}

Most traditional methods in the state-of-the-art are based on machine learning techniques with different feature extraction approaches such as HOG, LBP, and BSIF, or the fusion of some of them \cite{KUMARI2019}. However, today we have powerful pre-trained deep learning methods based on facial images. Using transfer learning techniques, the information extracted from some layers using fine-tuning techniques or embedding approaches could be suitable to perform periocular verification. This is the approach followed in this work.

This task involves information from periocular images estimating an eye embedding vector for a new given eye from a selfie image. An eye embedding is a vector that represents the features extracted from the eyes periocular images. This comparison occurs using euclidean distance to verify if the distance is below a predefined threshold, often tuned for a specific dataset or application.
For this paper, a VGGFace~\cite{Parkhi15}, FaceNet~\cite{facenet} and ArcFace ~\cite{deng2018arcface} models have been used as a feature extractor for periocular recognition. Also a comparison with BSIF handcrafted featured is included.

\section{Experimental Setup}
\label{experiments}

\subsection{Experimental Protocol}

In order to assess the soundness of the proposed method, we focus on a twofold objective: i) evaluate the SR approaches, and ii) analyse selfie periocular recognition systems using those SR techniques.

\textbf{Super-resolution models}. First, we have trained the DCSCN, WDSR-A, and SRGAN methods as a baseline for benchmarking purposes. The main properties and default parameters of those methods are summarised in the following:

\begin{itemize}
\item \textit{DCSCN}: Number of CNN layers = 12, Number of first CNN filters = 196, Number of last CNN filters = 48, Decay Gamma = 1.5, Self Ensemble = 8, Batch images for training epoch = 24,000, Dropout rate = 0.8, Optimiser function = Adam, Image size for each Batch = 48, Epochs = 100, Early stopping = 10.

\item \textit{WDSR-A}: Number of residual blocks = 8, Number of CNN layers in the main branch = 6, Number of expansion of residual blocks = 4, Number of filters main branch = 64, Number of filters residual blocks = 256, Activation function = Relu, Optimisation Function = Adam, Learning Rate = 1e-4 and 1-e-5, Beta = 1e-7, Size of batch images = 96, Number of steps = 60,000.

\item \textit{SRGAN}: The network has two modules:
\begin{itemize}
\item \textit{Generator}: This stage is used for learning the inverse function for downsampling the image and to generate the LR images from their corresponding HR, based in a pre-trained VGG-54. The following parameters are used: Number of residual blocks = 16, Number of CNN layers with residual blocks = 2, activation function residual block = PRelu, Kernel size residual block = 3, CNN layers = 3, kernel size = 9, 3 and, 9. Filters numbers = 64, Optimisation function = Adam, Learning rate = 1e-4 and 1e- 5, batch image size = 96, Steps = 100,000, mini size batches = 16.

\item \textit{Discriminator}: In order to evaluate the similarity between the images generated by the SR generator (VGG-54) and the HR images, the discriminator is trained with the following parameters: CNN layers = 8, Filter numbers:64, 64, 128, 128,  256, 256, 512 and 512. Kernel size = 3, activation function = Relu, Momentum batch normalisation = 0.8, Optimisation function = Adam, Learning Rate = 1e-5 and, 1e-6, Batch size = 16, Steps = 100,000.
\end{itemize}

\end{itemize}

Subsequently, we evaluated our ESISR method using the pixel-shuffle technique \cite{pixel_shuffle}. The best parameters for our approach were: Number of CNN layers = 7, Number of first CNN filters = 32, Number of last CNN filters = 8, Decay Gamma = 1.2, Self Ensemble = 8, Batch images for training epoch = 24,000, Dropout rate = 0.5, Optimiser function = Adam, Image size for each Batch = 32, Epochs= 100, Early stopping = 10. We further improved the efficiency of our proposal by using the transpose convolution instead of pixel-shuffle. 

In all experiments, we assess the quality of the produced SR images using the sharpness function defined in Eq.~\ref{eq:sharp}, and the efficiency in terms of the number of features and parameters. It should be noted that the True Sharpness represents the sharpness of the original image~((prior to downsampling), and Output Sharpness represents the sharpness of the reconstructed high resolution image created by ESISR. Therefore, the goal is to achieve an Output Sharpness as close as possible to the True Sharpness. From those experiments, we selected the configuration achieving the best performance. All methods were trained using the Samsung database and tested with the SET-5E dataset. 

\vspace*{0.2cm}
\textbf{Periocular SR verification}. We then extract the embedded information from selfie periocular images and compare the results with a handcrafted method for the periocular verification system. Afterwards, feature extraction was applied to the best super-resolved images using x2, x3, and x4 increased sizing, and it was compared with the same sizes but using traditional methods such as inter-area, lineal, and cubic. All the SR methods for periocular verification were tested using the MobBIO and NTNU datasets, which are different from the ones used to train the SR stage in order to grant unbiased results.

BSIF handcrafted features were used to extract textural information. An exhaustive exploration of the 60 filters was made. The image was divided into two rows and three columns. For each patch, a histogram was estimated. The concatenation of all the histograms represents the final vector. In this case, the $5\times5$-5, $9\times9$-5 and $11\times11$-5 bits show the best performances.

In more details, the FaceNet, VGGFace, and ArcFace pre-trained models were used to extract the embedding information. For FaceNet the feature vector has a size of 1,722 and input size image of $224 \times 224 \times  3$. For VGGFace, the feature vector has a size of 2,048 and input size image of $224 \times 224 \times  3$. ArcFace inputs have a size of 512 an input size of $112 \times 112 \times 3$.

A PC with Intel I7, 32 GB RAM, and GPU-1080TI was used for train all the stand-alone SR model.

\subsection{Databases}

\begin{figure}[t]
\centering
\includegraphics[width=0.9\linewidth]{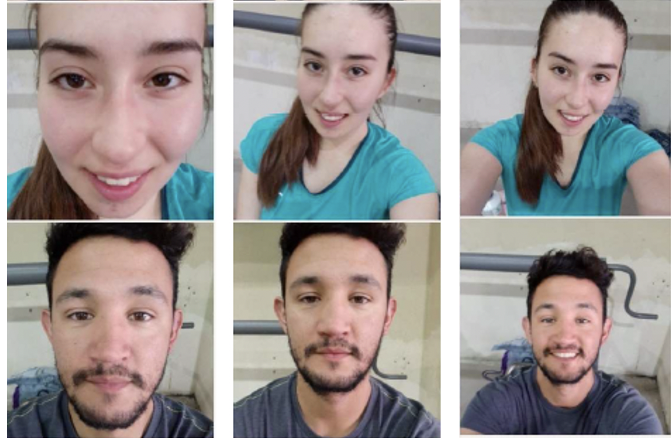}
\caption{\label{selfie} Example of Samsung databases. Left: closest position. Middle: half arm extended. Right: full arm extended.}
\end{figure}

In order to analyse the performance of the SR algorithm, four databases were used. A new dataset was acquired in a collaborative effort with subjects from different countries with Samsung smartphones using an app specially designed for this purpose: \url{visualselfie.org}\footnote{Only available from smartphones}. This app was designed in order to capture different variations of selfie scenarios in three distances, as depicted in Fig.~\ref{selfie}. More specifically, 800 images were selected to be used for training and 100 for testing\footnote{A similar number of images are used in the state-of-the-art for general-purpose methods; e.g. the DIV2K database}. From the training dataset, 228,700 patches of $48\times48$ px.\ were created for experiment 2 and $32\times32$ for experiment 3.

A second dataset, \textit{Set-5E}, was created to validate the results. This database has 100 images from different subjects acquired with different smartphones extracted from the CSIP database in the visual spectrum ~\cite{SANTOS}. It has 2004 images, stemming from 50 subjects over 10 different mobile setups.

\begin{figure}[h]
\centering
\includegraphics[width=0.24\linewidth]{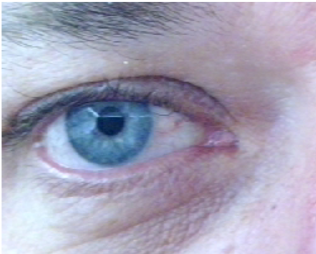}
\includegraphics[width=0.24\linewidth]{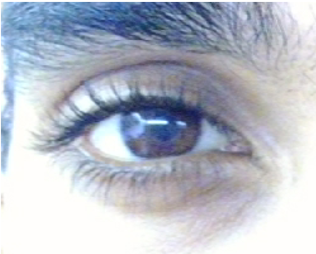}
\includegraphics[width=0.24\linewidth]{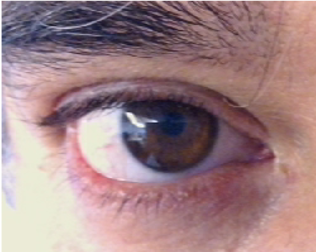}
\includegraphics[width=0.24\linewidth]{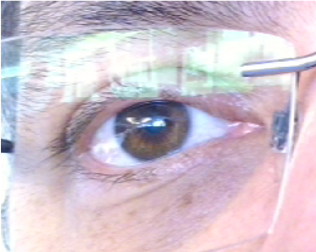}
\caption{MOBBIO database examples.}
\centering
\label{mobio}
\end{figure}

A third database MobBIO was used to super-resolved the size of the images with the best pre-trained super-resolution model~(ESISR). It was also used to measure the performance of the periocular verification system. The MobBIO dataset comprises the biometric data from 152 volunteers. Each subject provided samples of face, iris, and voice. There are on average 8 images for each subject from a NOKIA N93i mobile. Some examples are presented in Fig.~\ref{mobio}.

\begin{figure}[H]
\centering
\includegraphics[scale=0.40]{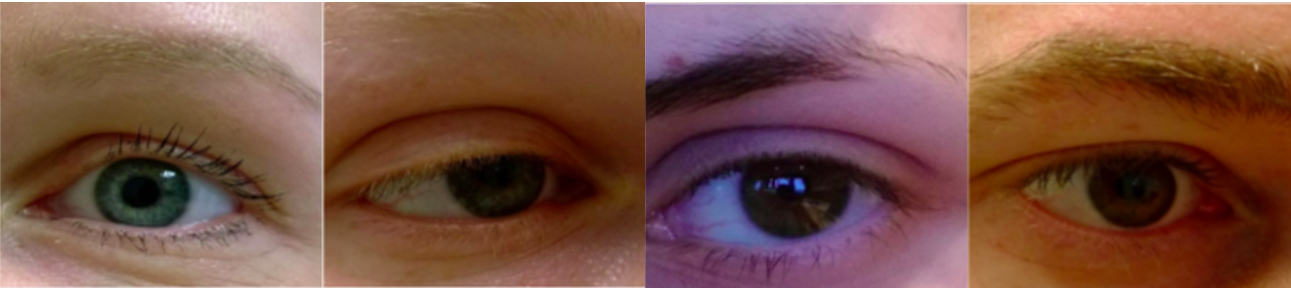}
\caption{NTNU database examples.}
\centering
\label{ntu_side}
\end{figure}

The last database is VISPI, captured by NTNU, which was used to measure the performance of the periocular verification system~\footnote{VISPI will be denoted as NTNU database}. The NTNU dataset comprises the biometric data from 152 volunteers and 3,139 total images. Each subject provided samples of left and right iris. There are in average 11 images for each subject from a NOKIA N93i mobile. Some examples are presented in Fig.~\ref{ntu_side}.

\section{Results and Discussion}
\label{sec:results}

\begin{table*}[]
\centering
\caption{Summary of the results for 3 different scales (x2, x3, and x4) for our system (ESISR) with different configurations and the benchmark with DCSCN, WDSR-A, and SRGAN. True Sharpness denotes the sharpness for the original image (LR), and Output Sharpness the sharpness for reconstructed SR images.}
\label{tab:table1}
\begin{tabular}{lcccccccc}
\toprule
\textbf{Method}  & \textbf{Conv.} & \textbf{\# Features} & \textbf{Scale} & \textbf{\# Param} & \textbf{PSNR} & \textbf{SSIM} & 
\textbf{True Sharp.} & \textbf{Output Sharp.} \\ \midrule
\multirow{3}{*}{DCSCN \cite{yamanaka}}& \multirow{3}{*}{} & \multirow{3}{*}{1,301} 
                   & x2    & 1,754,942   & 37.11. & 0.95 & 17.04 & 16.85           \\
&             &    & x3    & 2,170,142   & 32.82  & 0.91 & 18.05 & 16.45           \\
&             &    & x4    & 2,087,102   & 30.52  & 0.86 & 16.90 & 12.47                                      
\\ \midrule
\multirow{3}{*}{WDSR-A \cite{yu}} & \multirow{3}{*}{}&\multirow{3}{*}{} 
                    & x2   & 597,000 & 47.87& 0.98  & 17.04  & 10.89                    \\
& Pixel-shuffle &   & x3    & 603,000 & 46.59 & 0.97 & 18.05 & 10.82                  \\ 
&               &   & x4    & 610,000 & 43.92 & 0.94 & 16.90 & 10.72                                                 
\\ \midrule
\multirow{3}{*}{SRGAN \cite{ledig}}   & \multirow{3}{*}{} & \multirow{3}{*}{}
                    & x2 & 24.864.000   & 39.66 & 0.96 & 17.04 & 10.82                  \\
& Pixel-shuffle &   & x3 & 25.131.000 & 38.72 & 0.94  & 18.05 & 10.95                  \\ 
&               &   & x4 & 26.930.000 & 34.09 & 0.88  & 16.90 & 10.64                                              
\\ \midrule
\multirow{3}{*}{ESISR-1} & \multirow{3}{*}{Pixel-shuffle 48x48}& 
\multirow{3}{*}{1,000} &x2 & 27,209      & 36.49& 0.95  & 17.04& 16.70               \\ 
&               &    &x3 & 28,654      & 32.89& 0.90  & 18.05& 16.01               \\ 
&               &    &x4 & 64,201      & 29.08& 0.86  & 16.90& 12.00                                          
\\ \midrule
\multirow{3}{*}{ESISR-2} & \multirow{3}{*}{Pixel-shuffle 32x32}&
\multirow{3}{*}{131}   & x2& 27,209      & 38.91& 0.90  & 17.04& 15.43             \\ 
&               &    & x3& 28,654      & 36.78& 0.85  & 18.05& 15.46            \\ 
&               &    & x4& 64,201      & 35.47& 0.81  & 16.90& 16.34
\\ \midrule
\multirow{3}{*}{ESISR-3} & \multirow{3}{*}{Transpose Convolution} & 
\multirow{3}{*}{131}   & x2& 100,316     & 35.52& 0.81  & 17.04 & 14.38           \\ 
&               &    & x3& 109,564     & 36.84& 0.85  & 18.05 & 15.06          \\ 
&               &    & x4& 100,318     &35.52 & 0.81 & 16.90& 16.14
\\ \bottomrule
\end{tabular}

\end{table*}
 
\subsection{Super-resolution models}
\label{sec:SRmodels}

First, we establish a baseline by testing the DCSCN, WSDR-A, and SR-GAN models with their default parameters. Then, we analyse our proposal (ESISR-X) using \emph{pixel-shuffle} and the new loss function including the Sharpness metric (see Eqs.~\ref{eq:sharp} and~\ref{eq:loss}). Table~\ref{tab:table1} summarises the results: Rows 1-3 show the results for traditional SR methods (DSCN with 12 layers and $96\times96$ patches, WDSR-A with 8 residual blocks and $62\times62$ patches, SR-GAN with 16 residual blocks and $96\times96$ patches); and rows 4 to 6 present the results of our proposed method: ESISR-1 using the pixel-shuffle algorithm with only 7 convolutions layers and $48\times48$ patches, ESISR-2 using the pixel-shuffle algorithm with only 7 convolutions layers and $32\times32$ patches, and ESISR-3 using the transposed convolution algorithm with only 7 convolutions layers.  

Observing the results, we note that all the image enlargement x2, x3, and x4 extract the same number of features for each method (i.e., 1,301 for DCSNN and 1,000 for ESISR). The more considerable difference lies on the number of parameters of each method: while DCSCN, WSDR-A, and SR-GAN methods need a large number of parameters (for images increased by x2, 1,754,942, 597,000, and 24,864,000; for images increased by x3, 2,170,142, 603,000, and 25.131.000; and for images increased for x4, 2,087,102, 610,000, and 26,939,000), these numbers are drastically reduced by the our ESISR-1 proposed method, which needs only 27.209 parameters when the image is increased by x2, 28.654 parameters when increased by x3, and 64.201 parameters when increased by x4\footnote{Sample images are shown in the Appendix, Fig. \ref{fig:sr_example}}.

In addition to that gain in terms of efficiency, we may observe in Table~\ref{tab:table1} that the newly proposed loss function based on sharpness allows us to get a good reconstruction. The Output sharpness for each scale value is similar to the values obtained by DSCN (e.g. 16.85 \textit{vs}.\ 16.70 for x2), and also close to the target True Sharpness of 17.04. Therefore, we may conclude that the proposed method keeps the sharpness quality of the images, thereby making it suitable for SR applications for mobile devices.

In addition to the baseline configuration of ESISR-1, we also evaluated two additional approaches. First, the most efficient implementation of ESISR with a big reduction of features (down to 131) and a number of parameters with pixel-shuffle and $32\times32$ was analysed (Table~\ref{tab:table1}, row 5). Then, we also tested the method using \emph{transposed convolution} with the same number of 131 features (Table~\ref{tab:table1}, row 6). 
The Transpose convolutions layer is an inverse convolutions layer that will both up-sample input and learn how to fill in details during the model training process, at the cost of increasing the number of parameters (i.e., less efficient than pixel-shuffling). As we may observe in Table~\ref{tab:table1}, the pixel-shuffle with $32\times32$ px.\ uses the same number of parameters as with $48\times48$ px. In contrast, the transposed convolution requires 100,316 parameters when the image is increased by 2 (x2), 109,564 parameters when increased by 3 (x3), and 100,318 parameters when increased by 4 (x4). In spite of this increase, the ESISR is still 10 to 20 times more efficient than the traditional DCSCN.

Regarding the quality of the SR iris images, we can observe that both configurations tested in this last experiment (row 5-6) achieve a similar sharpness for the x3 and x4 scale values (14.43, 14.38, 15.46 and 16.32), but not for x2. In the latter case, the pixel-shuffle approach clearly outperforms the transpose-convolution method (15.43 \textit{vs}.\ 14.38). 
The lower result of reconstruction was reached for the SRGAN method with a higher number of parameters and a relevant difference of the value of output sharpness\footnote{Reconstruction examples are presented in the Appendix}.

\subsection{Periocular SR verification}

We now evaluate the periocular verification systems including the SR methods analysed in the previous section. In order to assess the quality of the super-resolved images, the MobBIO and NTNU datasets were used to evaluate the reconstruction performance with the best SR method proposed in Sect. \ref{sec:SRmodels}, namely ESISR with pixel-shuffle.

\begin{figure*}[]
\centering 
\includegraphics[width=0.43\linewidth]{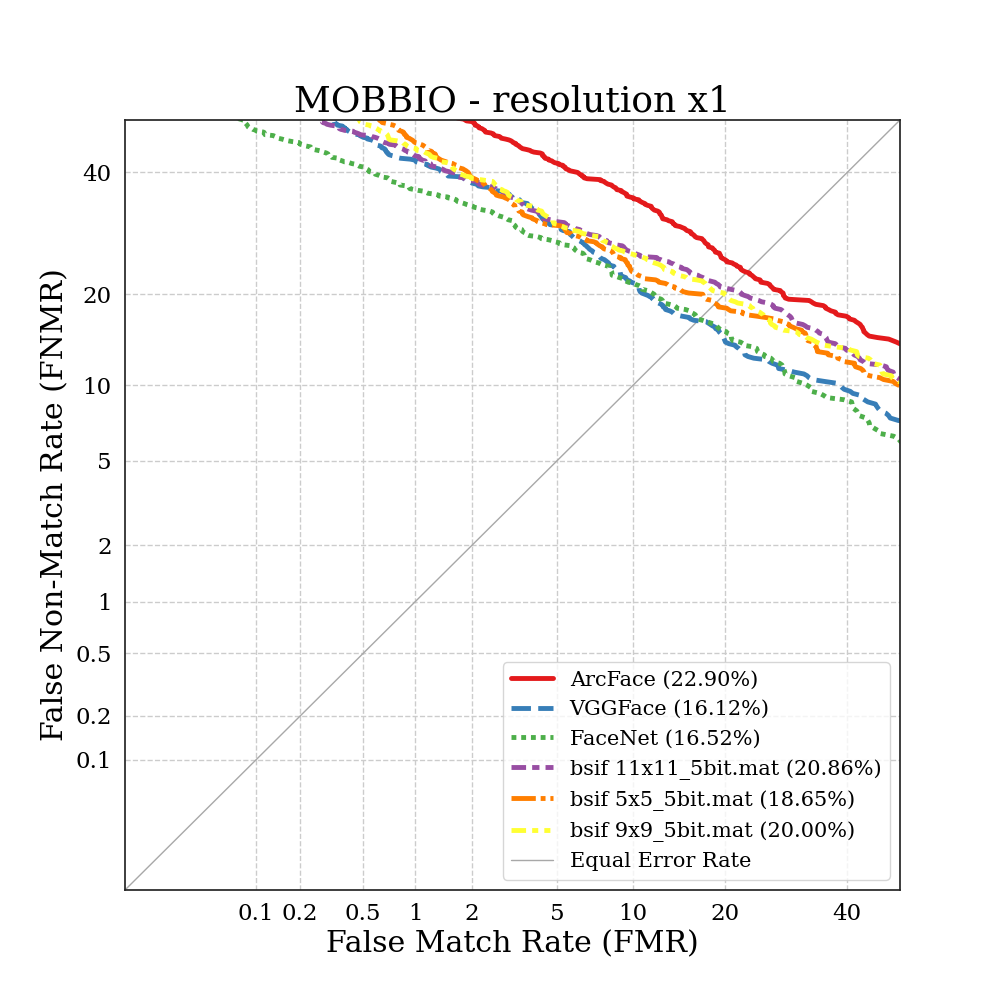}
\includegraphics[width=0.43\linewidth]{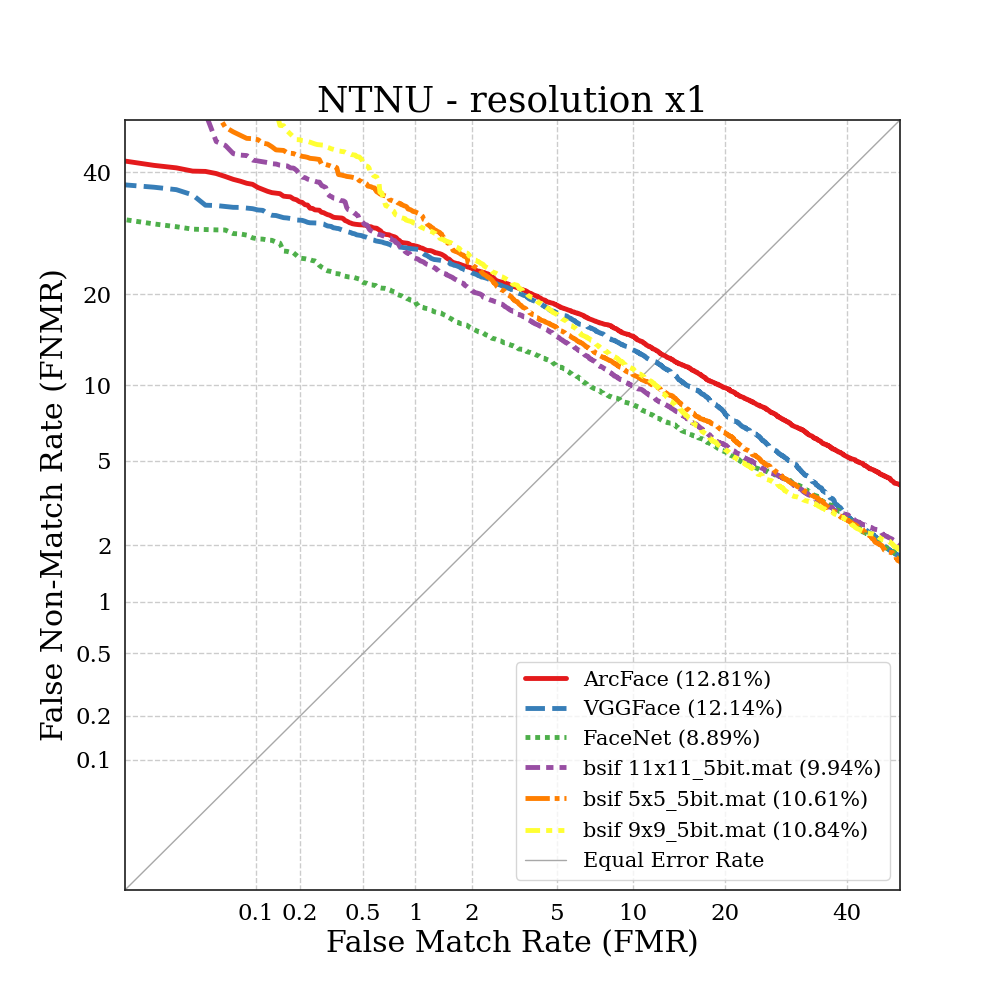}
\includegraphics[width=0.43\linewidth]{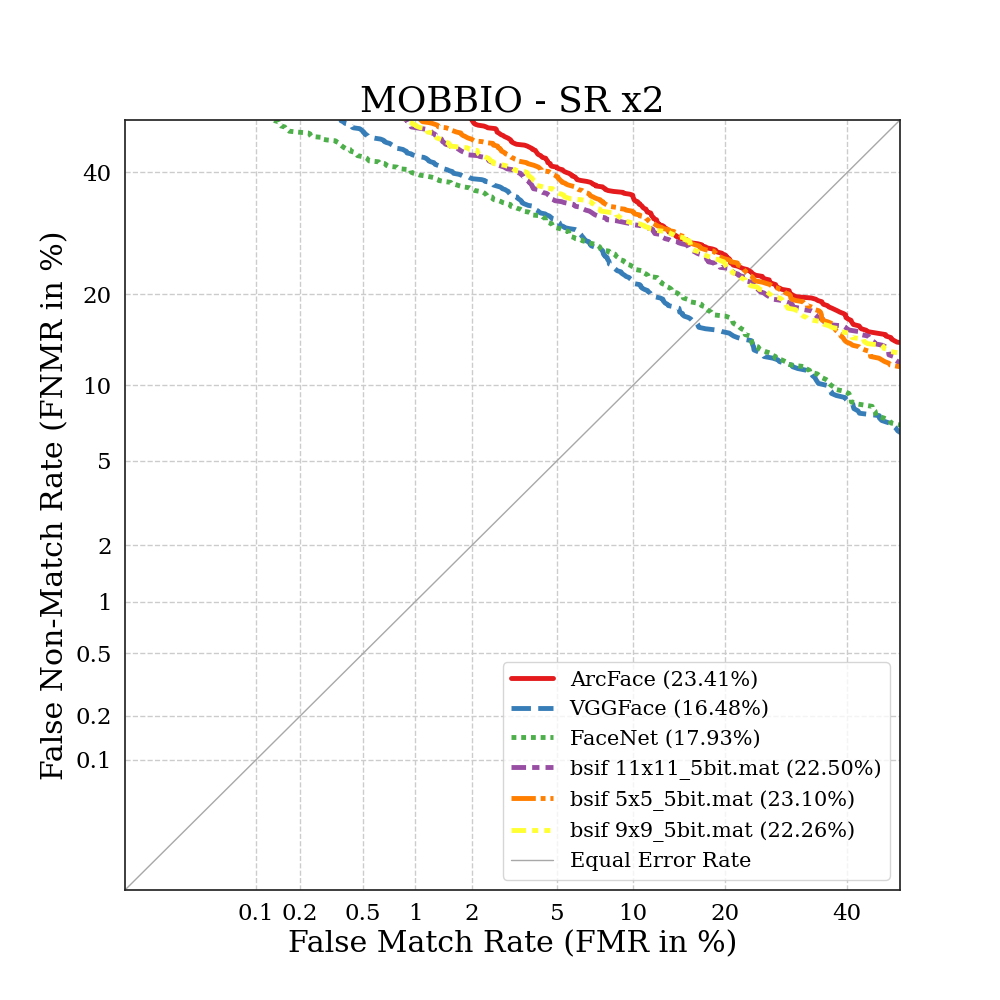}
\includegraphics[width=0.43\linewidth]{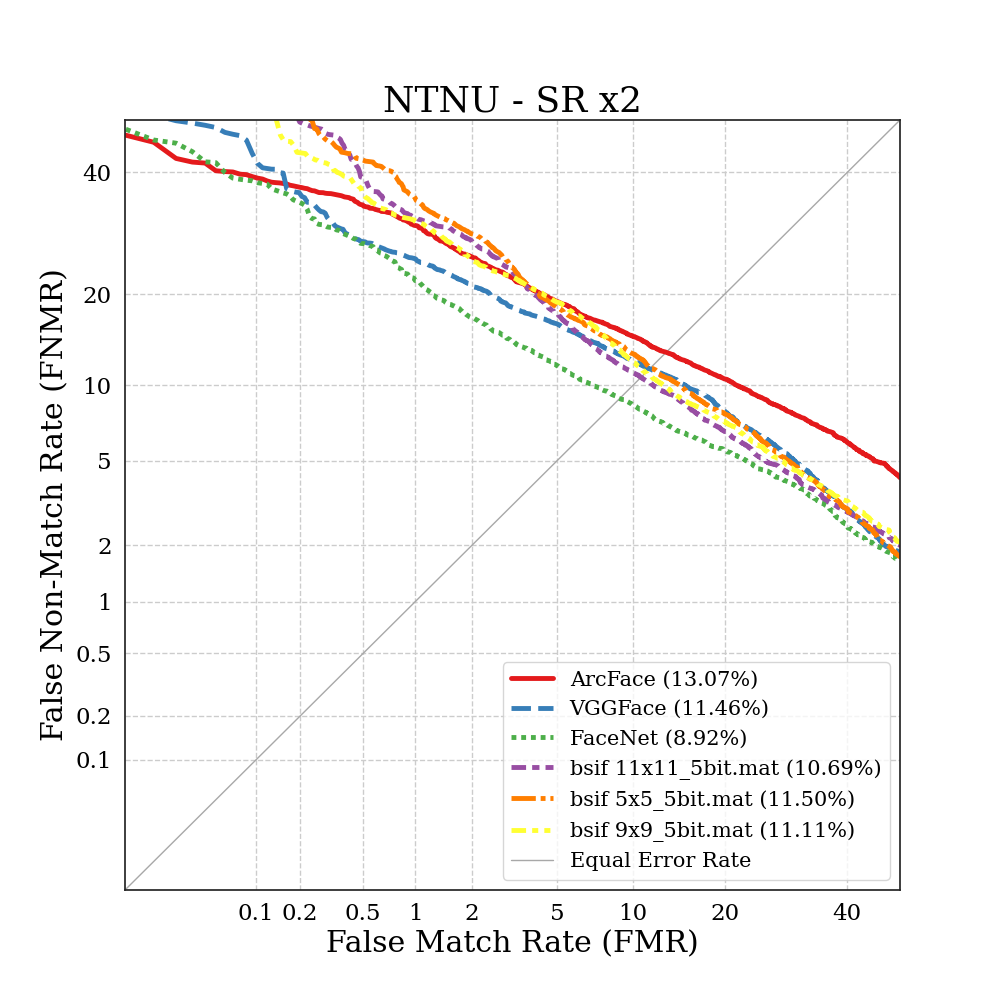}
\caption{DET curves for MobBIO and NTNU datasets using the SR method (x1 and x2) and including periocular recognition systems based on deep learning and handcrafted features (BSIF). The EER is shown in parenthesis for each technique.} 
\label{DETs-sr-mobio-ntu1}
\end{figure*}

\begin{figure*}[]
\centering 
\includegraphics[width=0.45\linewidth]{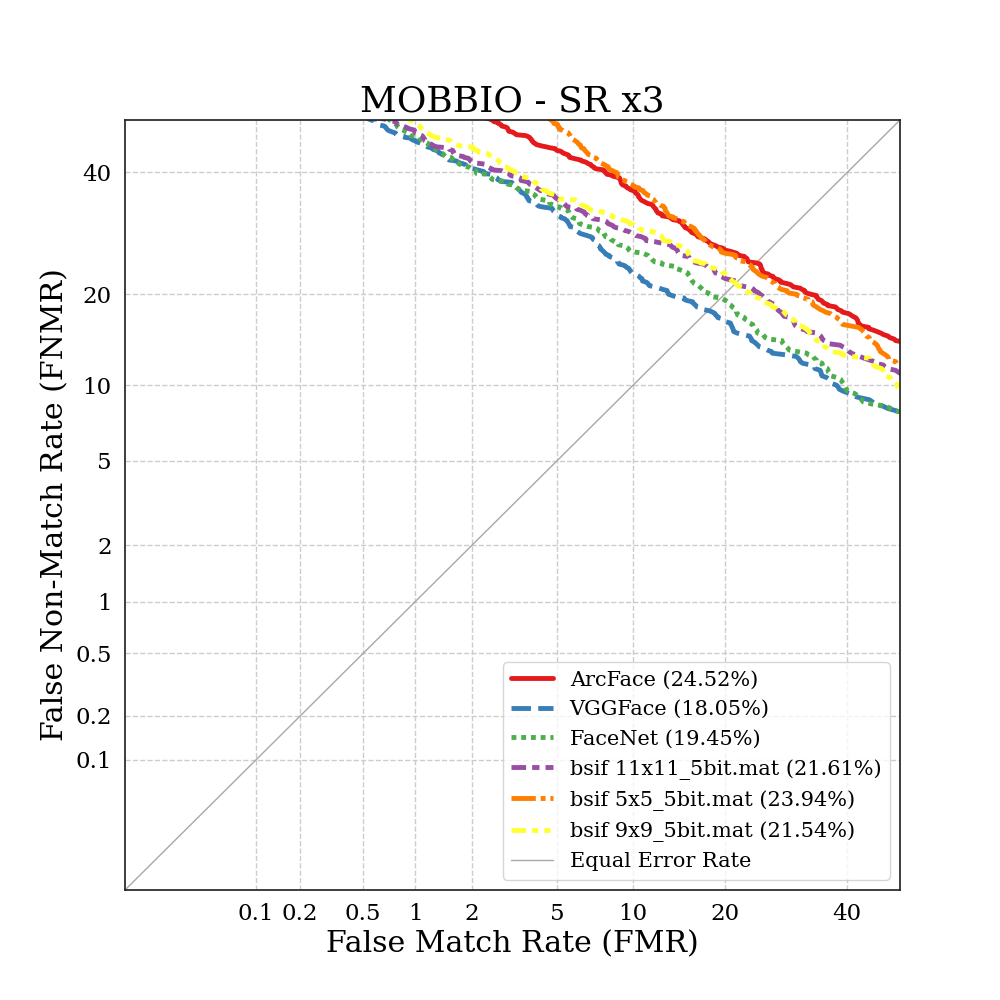}
\includegraphics[width=0.45\linewidth]{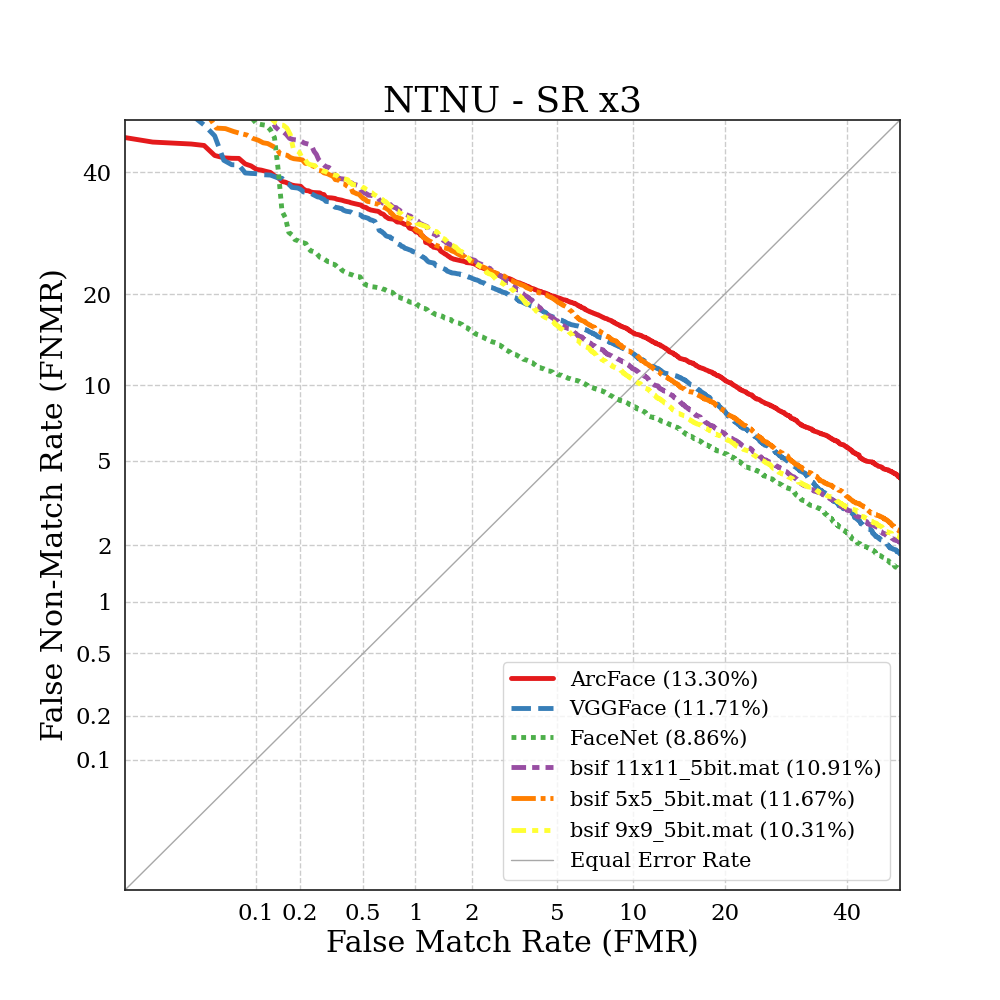}
\includegraphics[width=0.45\linewidth]{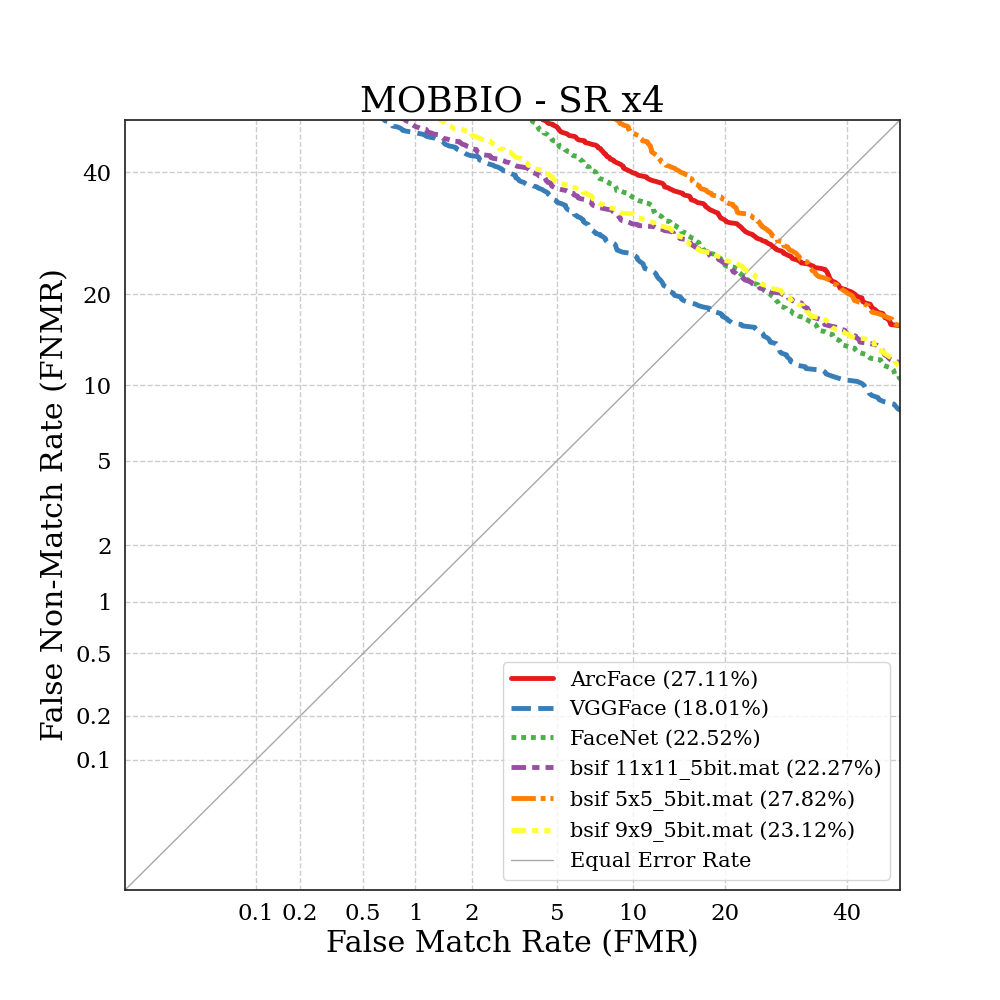}
\includegraphics[width=0.45\linewidth]{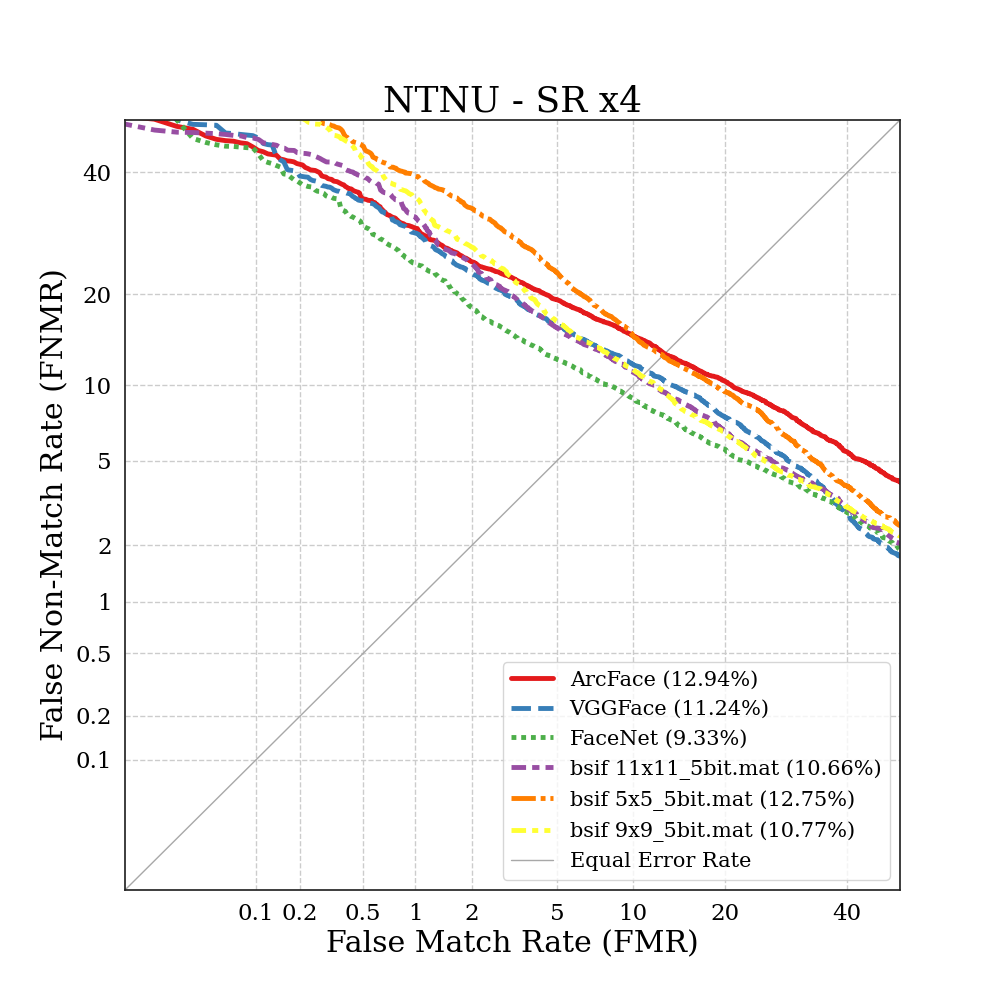}
\caption{DET curves for MobBIO and NTNU datasets using the SR method (x3 and x4) and including periocular recognition systems based on deep learning and handcrafted features (BSIF). The EER is shown in parenthesis.}. 
\label{DETs-sr-mobio-ntu2}
\end{figure*}

Figs.~\ref{DETs-sr-mobio-ntu1} and \ref{DETs-sr-mobio-ntu2} show the DET curves of the periocular verification system for MobBIO and NTNU datasets with a standard resolution (Resolution x1) in comparison with SR images resized by x2, x3, and x4 using the ESISR method. The results show VGGFace, FaceNet, ArcFace and three different BSIFs filters with equal error rates for each one.
An essential fact that we can see in Figures ~\ref{DETs-sr-mobio-ntu1} and \ref{DETs-sr-mobio-ntu2}, in this case, is that SR methods help maintain the recognition accuracy when selfies are captured at different distances instead of improving the eye recognition performance.

For images with a standard resolution (Resolution x1), VGGFace obtained the best results with an EER of 16.12\% using the MobBIO dataset. The best results were obtained for images from NTNU FaceNet with an EER of 8.89\%.

For images with an SR x2, x3 and x4, FaceNet outperforms VGGFace and ArcFace. Obtained an 8.92\%, 8.86\% and 9.33\% for the NTNU dataset. Conversely, MobBIO reached the lowest results. SR methods x2, x3 and x4 yielded 17.93\%, 19.45\% and 22.52\% respectively. Regarding BSIF, the three proposed filters reached a lower performance with EERs over 20\%. ArcFace obtained the worse results for the deep learning method for both datasets.

It is essential to highlight that the three scales keep the periocular verification quality based on the proposed perceptual sharpness loss. Thus, a weighted perceptual loss help to keep the quality of the images based on Sharpness metrics. This metric is more suitable for applying periocular iris images with SR than the traditional SNR and SSIM.

Table~\ref{tab:Results_mobio_ntnu} (top) shows the results for MobBIO Dataset and present different sizes of SR images increased by a factor of x2, x3, and x4 and its benchmark with the pre-trained FaceNet VGGFace, ArcFace and BSIF filters as a feature extractor. Also, three resized operations were explored, analysed and compared when used in super-resolution techniques, Inter-lineal, Inter-cubic, or Inter-area resized. The results reach slightly change when used on the super-resolution process\footnote{A benchmark with traditional resizing methods such as InterArea, InterCubic and InterLineal was performed DET curves are shown in the Appendix}. We can observe that the features extracted from Deep learning methods (embeddings) performed better than BSIF filters. Overall, FaceNet reached the best results in all the models with x2, x3 and x4 in comparison with VGGFace in NTNU dataset.  This result is interesting for high-security applications since operating points are usually defined at small FMR values. 

The results are related to the size of the embedded vector extracted from the pre-trained model. The features extracted from FaceNet are more representative and general-purpose than for VGGFace and ArcFace.

Table~\ref{tab:Results_mobio_ntnu} (bottom) shows the results for NTNU Dataset and present different sizes of SR images increased by a factor of x2, x3, and x4 and its benchmark with the pre-trained FaceNet VGGFace, ArcFace and BSIF filters as a feature extractor. In addition, a comparison with traditional resize methods such as InterArea, InterCubic and InterLineal were performed. Column one shows the name of all techniques explored. Column  4 up to column 6 show the results of the best BSIF filters selected. The results reported show the EER and False Not Match Rate (FNMR) based on False Match Rate (FMR) at 10\%.

\begin{table*}[]
\centering
\caption{MobBIO (top) and NTNU (bottom) Verification results with No resizing (resolution x1), interArea, interCubic, and interLineal. Both EER and FNMR are presented in \%, and FNMR is given at FMR = 10\%.}
\label{tab:Results_mobio_ntnu}
\begin{tabular}{|ccccccccccccc|}
\toprule
\multicolumn{13}{c}{MobBIO Dataset}                                                     \\ \midrule
\multicolumn{1}{c}{}                & \multicolumn{2}{|c|}{FaceNet}                                                                         & \multicolumn{2}{c|}{VGGFace}                                                                         & \multicolumn{2}{c|}{ArcFace}                                                                         & \multicolumn{2}{c|}{\begin{tabular}[c]{@{}c@{}}BSIF\\ 5x5\_5bits\end{tabular}}                       & \multicolumn{2}{c|}{\begin{tabular}[c]{@{}c@{}}BSIF\\ 9x9\_5bits\end{tabular}}                       & \multicolumn{2}{c}{\begin{tabular}[c]{@{}c@{}}BSIF\\ 11x11\_5bits\end{tabular}}                     \\ \midrule
\multicolumn{1}{l}{SR Method}          & \multicolumn{1}{|c}{\begin{tabular}[c]{@{}c@{}}EER\end{tabular}} & \multicolumn{1}{c|}{FNMR}  & \multicolumn{1}{c}{\begin{tabular}[c]{@{}c@{}}EER\end{tabular}} & \multicolumn{1}{c|}{FNMR}  & \multicolumn{1}{c}{\begin{tabular}[c]{@{}c@{}}EER\end{tabular}} & \multicolumn{1}{c|}{FNMR}  & \multicolumn{1}{c}{\begin{tabular}[c]{@{}c@{}}EER\end{tabular}} & \multicolumn{1}{c|}{FNMR}  & \multicolumn{1}{c}{\begin{tabular}[c]{@{}c@{}}EER\end{tabular}} & \multicolumn{1}{c|}{FNMR}  & \multicolumn{1}{c}{\begin{tabular}[c]{@{}c@{}}EER\end{tabular}} & \multicolumn{1}{c}{FNMR}                \\ \midrule
\multicolumn{1}{l|}{\textbf{No Redimension}}  & \multicolumn{1}{c}{\textbf{16.52}}                                              & \multicolumn{1}{c|}{16.83} & \multicolumn{1}{c}{\textbf{16.12}}                                              & \multicolumn{1}{c|}{16.67} & \multicolumn{1}{c}{\textbf{22.90}}                                              & \multicolumn{1}{c|}{23.16} & \multicolumn{1}{c}{\textbf{18.65}}                                              & \multicolumn{1}{c|}{20.50} & \multicolumn{1}{c}{\textbf{20.00}}                                              & \multicolumn{1}{c|}{20.50} & \multicolumn{1}{c}{\textbf{20.86}}                                              & \multicolumn{1}{c}{21.00} \\ \midrule
\multicolumn{1}{l|}{Inter-Area x2}   & \multicolumn{1}{c}{16.17}                                              & \multicolumn{1}{c|}{16.16} & \multicolumn{1}{c}{16.00}                                              & \multicolumn{1}{c|}{15.83} & \multicolumn{1}{c}{22.62}                                              & \multicolumn{1}{c|}{\textbf{22.50}} & \multicolumn{1}{c}{20.72}                                              & \multicolumn{1}{c|}{20.66} & \multicolumn{1}{c}{20.68}                                              & \multicolumn{1}{c|}{20.83} & \multicolumn{1}{c}{21.57}                                              & \multicolumn{1}{c}{21.33}                      \\
\multicolumn{1}{l|}{Inter-Cubic x2}  & \multicolumn{1}{c}{16.00}                                              & \multicolumn{1}{c|}{16.16} & \multicolumn{1}{c}{16.33}                                              & \multicolumn{1}{c|}{15.83} & \multicolumn{1}{c}{22.12}                                              & \multicolumn{1}{c|}{22.00} & \multicolumn{1}{c}{20.97}                                              & \multicolumn{1}{c|}{20.66} & \multicolumn{1}{c}{19.86}                                              & \multicolumn{1}{c|}{20.83} & \multicolumn{1}{c}{21.33}                                              & \multicolumn{1}{c}{21.33}                       \\ 
\multicolumn{1}{l|}{Inter-Lineal x2} & \multicolumn{1}{c}{17.71}                                              & \multicolumn{1}{c|}{17.16} & \multicolumn{1}{c}{16.00}                                              & \multicolumn{1}{c|}{15.83} & \multicolumn{1}{c}{23.67}                                              & \multicolumn{1}{c|}{23.66} & \multicolumn{1}{c}{22.79}                                              & \multicolumn{1}{c|}{22.83} & \multicolumn{1}{c}{20.87}                                              & \multicolumn{1}{c|}{21.16} & \multicolumn{1}{c}{21.56}                                              & \multicolumn{1}{c}{21.33}                      \\ 
\multicolumn{1}{l|}{\textbf{ESISR x2}}        & \multicolumn{1}{c}{\textbf{17.93}}                                              & \multicolumn{1}{c|}{17.83} & \multicolumn{1}{c}{\textbf{16.48}}                                              & \multicolumn{1}{c|}{16.33} & \multicolumn{1}{c}{\textbf{23.41}}                                              & \multicolumn{1}{c|}{23.33} & \multicolumn{1}{c}{\textbf{23.10}}                                              & \multicolumn{1}{c|}{23.00} & \multicolumn{1}{c}{\textbf{22.26}}                                              & \multicolumn{1}{c|}{22.33} & \multicolumn{1}{c}{22.50}                                              & \multicolumn{1}{c}{22.50}                       \\ \midrule
\multicolumn{1}{l|}{Inter-Area x3}   & \multicolumn{1}{c}{20.67}                                              & \multicolumn{1}{c|}{19.66} & \multicolumn{1}{c}{15.67}                                              & \multicolumn{1}{c|}{15.66} & \multicolumn{1}{c}{24.75}                                              & \multicolumn{1}{c|}{24.83} & \multicolumn{1}{c}{28.68}                                              & \multicolumn{1}{c|}{28.66} & \multicolumn{1}{c}{20.65}                                              & \multicolumn{1}{c|}{20.83} & \multicolumn{1}{c}{21.41}                                              & \multicolumn{1}{c}{21.50}                       \\ 
\multicolumn{1}{l|}{Inter-Cubic x3}  & \multicolumn{1}{c}{17.67}                                              & \multicolumn{1}{c|}{18.16} & \multicolumn{1}{c}{16.00}                                              & \multicolumn{1}{c|}{16.00} & \multicolumn{1}{c}{24.12}                                              & \multicolumn{1}{c|}{24.16} & \multicolumn{1}{c}{23.97}                                              & \multicolumn{1}{c|}{23.50} & \multicolumn{1}{c}{21.07}                                              & \multicolumn{1}{c|}{20.66} & \multicolumn{1}{c}{22.04}                                              & \multicolumn{1}{c}{22.00}                      \\ 
\multicolumn{1}{l|}{Inter-Lineal x3} & \multicolumn{1}{c}{19.00}                                              & \multicolumn{1}{c|}{18.66} & \multicolumn{1}{c}{16.00}                                              & \multicolumn{1}{c|}{15.50} & \multicolumn{1}{c}{25.91}                                              & \multicolumn{1}{c|}{25.83} & \multicolumn{1}{c}{21.19}                                              & \multicolumn{1}{c|}{26.66} & \multicolumn{1}{c}{26.65}                                              & \multicolumn{1}{c|}{21.16} & \multicolumn{1}{c}{22.36}                                              & \multicolumn{1}{c}{22.50}                      \\ 
\multicolumn{1}{l|}{\textbf{ESISR x3}}        & \multicolumn{1}{c}{\textbf{19.45}}                                              & \multicolumn{1}{c|}{19.50} & \multicolumn{1}{c}{\textbf{18.05}}                                              & \multicolumn{1}{c|}{17.87} & \multicolumn{1}{c}{\textbf{24.52}}                                              & \multicolumn{1}{c|}{24.50} & \multicolumn{1}{c}{\textbf{23.94}}                                              & \multicolumn{1}{c|}{23.83} & \multicolumn{1}{c}{\textbf{21.54}}                                              & \multicolumn{1}{c|}{21.50} & \multicolumn{1}{c}{\textbf{21.61}}                                              & \multicolumn{1}{c}{21.83}                     \\ \midrule
\multicolumn{1}{l|}{Inter Area x4}   & \multicolumn{1}{c}{27.00}                                              & \multicolumn{1}{c|}{27.66} & \multicolumn{1}{c}{19.50}                                              & \multicolumn{1}{c|}{19.83} & \multicolumn{1}{c}{27.65}                                              & \multicolumn{1}{c|}{27.50} & \multicolumn{1}{c}{43.48}                                              & \multicolumn{1}{c|}{43.33} & \multicolumn{1}{c}{21.77}                                              & \multicolumn{1}{c|}{21.83} & \multicolumn{1}{c}{22.41}                                              & \multicolumn{1}{c}{22.50}                       \\ 
\multicolumn{1}{l|}{Inter-Cubic x4}  & \multicolumn{1}{c}{22.00}                                              & \multicolumn{1}{c|}{21.50} & \multicolumn{1}{c}{19.00}                                              & \multicolumn{1}{c|}{18.66} & \multicolumn{1}{c}{27.67}                                              & \multicolumn{1}{c|}{27.16} & \multicolumn{1}{c}{27.34}                                              & \multicolumn{1}{c|}{27.33} & \multicolumn{1}{c}{21.31}                                              & \multicolumn{1}{c|}{21.66} & \multicolumn{1}{c}{22.31}                                              & \multicolumn{1}{c}{22.67}                      \\ 
\multicolumn{1}{l|}{Inter-Lineal x4} & \multicolumn{1}{c}{23.00}                                              & \multicolumn{1}{c|}{23.00} & \multicolumn{1}{c}{18.00}                                              & \multicolumn{1}{c|}{18.33} & \multicolumn{1}{c}{28.06}                                              & \multicolumn{1}{c|}{28.33} & \multicolumn{1}{c}{29.59}                                              & \multicolumn{1}{c|}{29.83} & \multicolumn{1}{c}{23.22}                                              & \multicolumn{1}{c|}{23.16} & \multicolumn{1}{c}{23.13}                                              & \multicolumn{1}{c}{23.00}                       \\ 
\multicolumn{1}{l|}{\textbf{ESISR x4}}        & \multicolumn{1}{c}{\textbf{22.52}}                                              & \multicolumn{1}{c|}{22.50} & \multicolumn{1}{c}{\textbf{18.01}}                                              & \multicolumn{1}{c|}{18.00} & \multicolumn{1}{c}{\textbf{27.11}}                                              & \multicolumn{1}{c|}{27.00} & \multicolumn{1}{c}{\textbf{27.82}}                                              & \multicolumn{1}{c|}{28.00} & \multicolumn{1}{c}{\textbf{23.12}}                                              & \multicolumn{1}{c|}{23.33} & \multicolumn{1}{c}{\textbf{22.27}}                                              & \multicolumn{1}{c}{22.50}                       \\ \bottomrule
\end{tabular}

\vspace*{0.5cm}
\begin{tabular}{l|cc|cc|cc|cc|cc|cc}
\toprule
\multicolumn{13}{c}{NTNU Dataset}                                                     \\ \midrule
                        & \multicolumn{2}{c|}{FaceNet} & \multicolumn{2}{c|}{VGGFace} & \multicolumn{2}{c|}{ArcFace} & \multicolumn{2}{c|}{\begin{tabular}[c]{@{}c@{}}BSIF\\ 5x5\_5bits\end{tabular}} & \multicolumn{2}{c|}{\begin{tabular}[c]{@{}c@{}}BSIF\\ 9x9\_5bits\end{tabular}} & \multicolumn{2}{c}{\begin{tabular}[c]{@{}c@{}}BSIF\\ 11x11\_5bits\end{tabular}} \\ \midrule
SR Method                  & EER               & FNMR     & EER               & FNMR     & EER               & FNMR     & EER                                        & FNMR                              & EER                                        & FNMR                              & EER                               & FNMR                                        \\ \midrule
\textbf{No Redimension} & \textbf{8.89}              & 8.88     & \textbf{12.14}             & 12.10    & \textbf{12.81}             & 12.79    & \textbf{10.61}                                      & 10.60                             & \textbf{10.84}                                      & 10.92                             & \textbf{9.94 }                            & \multicolumn{1}{l}{10.12}                   \\ \midrule
Inter-Area x2           & 8.55              & 8.52     & 12.42             & 12.38    & 12.36             & 12.40    & 10.46                                      & 10.47                             & 9.76                                       & 9.77                              & 10.73                             & 10.74                                       \\
Inter-Cubic x2          & 8.52              & 8.53     & 11.66             & 11.64    & 12.49             & 12.46    & 10.49                                      & 10.49                             & 9.91                                       & 9.92                              & 10.77                             & 10.78                                       \\
Inter-Lineal x2         & 8.74              & 8.74     & 12.39             & 12.28    & 12.44             & 12.43    & 11.00                                      & 11.00                             & 9.82                                       & 9.83                              & 10.71                             & 10.67                                       \\
\textbf{ESISR x2}       & \textbf{8.92}     & 8.91     & \textbf{11.46}    & 11.47    & \textbf{13.07}    & 13.10    & \textbf{11.50}                             & 11.54                             & \textbf{11.11}                             & 11.11                             & \textbf{10.69}                    & 10.70                                       \\ \midrule
Inter-Area x3           & 12.06             & 12.07    & 12.51             & 12.51    & 12.94             & 12.93    & 12.29                                      & 12.75                             & 12.69                                      & 12.70                             & 11.72                             & 11.72                                       \\
Inter-Cubic x3          & 12.49             & 12.38    & 12.91             & 12.90    & 13.03             & 13.04    & 12.17                                      & 12.28                             & 12.85                                      & 12.84                             & 12.09                             & 12.12                                       \\
Inter-Lineal x3         & 12.86             & 12.86    & 12.49             & 12.48    & 12.88             & 12.87    & 12.19                                      & 21.19                             & 12.30                                      & 12.28                             & 11.28                             & 11.27                                       \\
\textbf{ESISR x3}       & \textbf{8.85}     & 8.84     & \textbf{11.71}    & 11.70    & \textbf{13.30}    & 13.29    & \textbf{11.67}                             & 11.66                             & \textbf{10.30}                             & 10.29                             & \textbf{10.91}                    & 10.89                                       \\ \midrule
Inter Area x4           & 16.59             & 16.58    & 13.45             & 12.51    & 14.51             & 14.54    & 13.25                                      & 13.24                             & 12.05                                      & 12.06                             & 11.51                             & 11.52                                       \\
Inter-Cubic x4          & 18.31             & 18.30    & 14.68             & 14.66    & 14.49             & 14.47    & 13.94                                      & 13.94                             & 12.60                                      & 12.60                             & 12.06                             & 12.06                                       \\
Inter-Lineal x4         & 18.94             & 16.92    & 13.50             & 13.51    & 14.57             & 14.57    & 15.18                                      & 15.16                             & 11.90                                      & 11.89                             & 10.99                             & 10.98                                       \\
\textbf{ESISR x4}       & \textbf{9.32}     & 9.32     & \textbf{11.24}    & 11.23    & \textbf{12.93}    & 12.92    & \textbf{12.75}                             & 12.75                             & \textbf{10.77}                             & 10.77                             & \textbf{10.66}                    & 10.65                                       \\ \bottomrule
\end{tabular}
\end{table*}


\section{Conclusion}
\label{conclusions}

In this paper, we have proposed an efficient and accurate image super resolution method focused on the generation of enhanced eyes images for periocular verification purposes using selfie images. To that end, we developed a two-stage approach based on a CNN with pixel-shuffle, a new loss function based on a sharpness metric (see Eq.~\ref{eq:sharp}), derived from the ISO/IEC 29794-6 standard for iris quality, and a selfie periocular verification proposal.

In the feature extraction stage of our method, the structure of the CNN model extracts optimised features, which are subsequently sent to the reconstruction network. In this latter network, we only used a recursive convolutional block with pixel-shuffle to obtain a better reconstruction performance with reduced computational requirements. In addition, the model is designed to be capable of processing original size images. Using these techniques, our model can achieve state-of-the-art performance with a fewer number of parameters (from the state-of-the-art DSCN with 2 million parameters, we achieve a comparable quality with 27,000 parameters). 

The perceptual loss function based on image sharpness that we propose allows us to keep the sharpness of iris images in the reconstructed images by x2, x3, and x4. This approach to improving the quality of the reconstruction and the SR in periocular recognition systems is well suited for implementation in mobile devices.

Regarding periocular verification system, as expected, the deep learning method's yielded better results than handcrafted methods. FaceNet achieved the best results in comparison to VGGFace and ArcFace. An EER of 8.7\%  without SR and 9.2\% for x2, 8.9\% for x3, and 9.5\% for x4 was obtained, respectively. Conversely, a slight performance was reached when VGGFace was used. An EER of 10.05\%  without SR and 9.94\%  for x2, 9.92\%  for x3, and 9.90\%  for x4, respectively.

Overall, there are marginal improvements for verification systems when only the size of the images is considered in combination with SR images. The information extracted with an embedded vector from the periocular area with a pre-trained model has a high quality of data for verification than BSIF because of the huge number of filters used during the training process. 

The uncontrolled conditions such as sunlight, occlusions, rotations, or the number of people in an image when a remote selfie is captured could be more challenging than the image size for RGB selfie images. This improvement to NIR iris images must be studied in a separate work. Those uncontrolled conditions need to be examined to improve the selfie periocular verification systems. 

In this research, SR helps maintaining the recognition accuracy when selfies are captured at different distances. That is, in realistic scenarios in contrast to fully controlled conditions. Our system was tested on images acquired at three different distances and obtained similar results to a baseline system with a unique acquisition distance, even when the selfie was resize using SR with x2, x3 and x4.

In future work, we will continue to collect images to train a specific periocular verification system based on CNN from scratch and/or using transfer-domain techniques. Concerning the number of images, we believe that if we use state-of-the-art pre-trained models, the machine learning-based methods could be replaced by the CNN models. The selection of the pre-trained models should be taken into account.

\section*{Acknowledgements}

European Union’s Horizon 2020 research and innovation program under grant agreement	883356, the DFG-ANR RESPECT Project (406880674), and the German Federal Ministry of Education and Research and the Hessian Ministry of Higher Education, Research, Science and the Arts within their joint support of the National Research Center for Applied Cybersecurity ATHENE.

\bibliographystyle{IEEEtran}
\bibliography{references.bib}
\vspace{-0.3cm}
\begin{IEEEbiography}[{\includegraphics[width=1in,height=1.25in,clip,keepaspectratio]{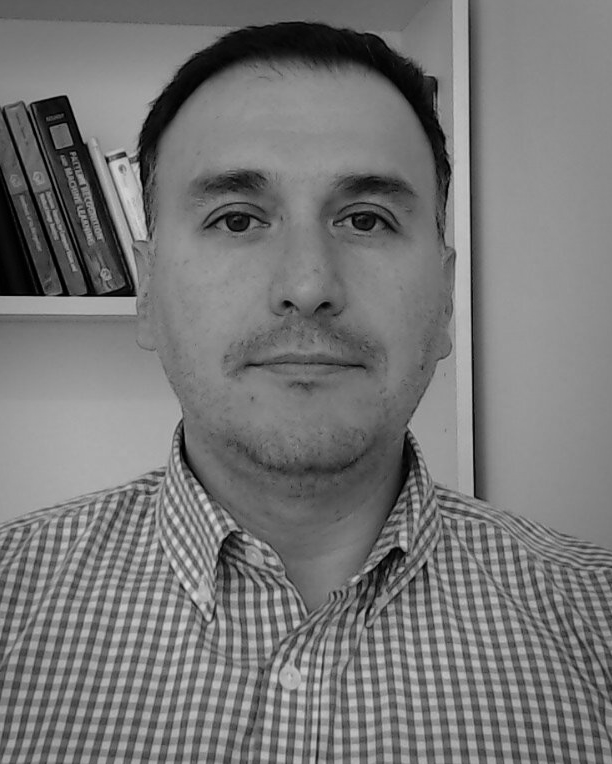}}]{Juan Tapia} received a P.E. degree in Electronics Engineering from Universidad Mayor in 2004, a M.S. in Electrical Engineering from Universidad de Chile in 2012, and a Ph.D. from the Department of Electrical Engineering, Universidad de Chile in 2016. In addition, he spent one year of internship at University of Notre Dame. In 2016, he received the award for best Ph.D. thesis. From 2016 to 2017, he was an Assistant Professor at Universidad Andres Bello. From 2018 to 2020, he was the R\&D Director for the area of Electricity and Electronics at Universidad Tecnologica de Chile - INACAP. He is currently a Senior Researcher at Hochschule Darmstadt~(HDA), and R\&D Director of TOC Biometrics. His main research interests include pattern recognition and deep learning applied to iris biometrics, morphing, feature fusion, and feature selection. 
\end{IEEEbiography}
\vspace{-0.3cm}

\begin{IEEEbiography}[{\includegraphics[width=1in,height=1.25in,clip,keepaspectratio]{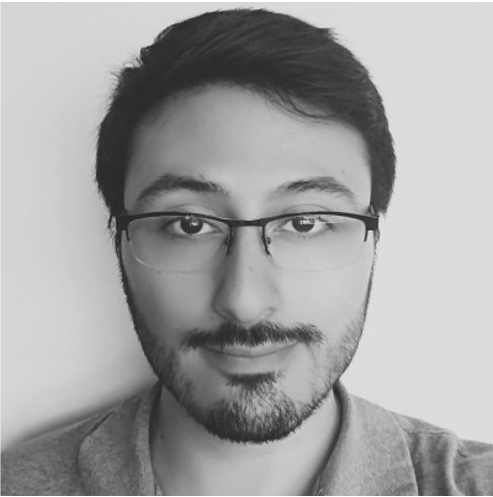}}]{Andres Valenzuela} received a B.S. in Computer Engineering from Universidad Andres Bello in 2019. Currently, he is a researcher at TOC Biometrics company. His main interests include computer vision, pattern recognition and deep learning applied to real problems such as tampering detection, classification and segmentation.
\end{IEEEbiography}

\vspace{-0.3cm}
\begin{IEEEbiography}[{\includegraphics[width=1in,height=1.25in,clip,keepaspectratio]{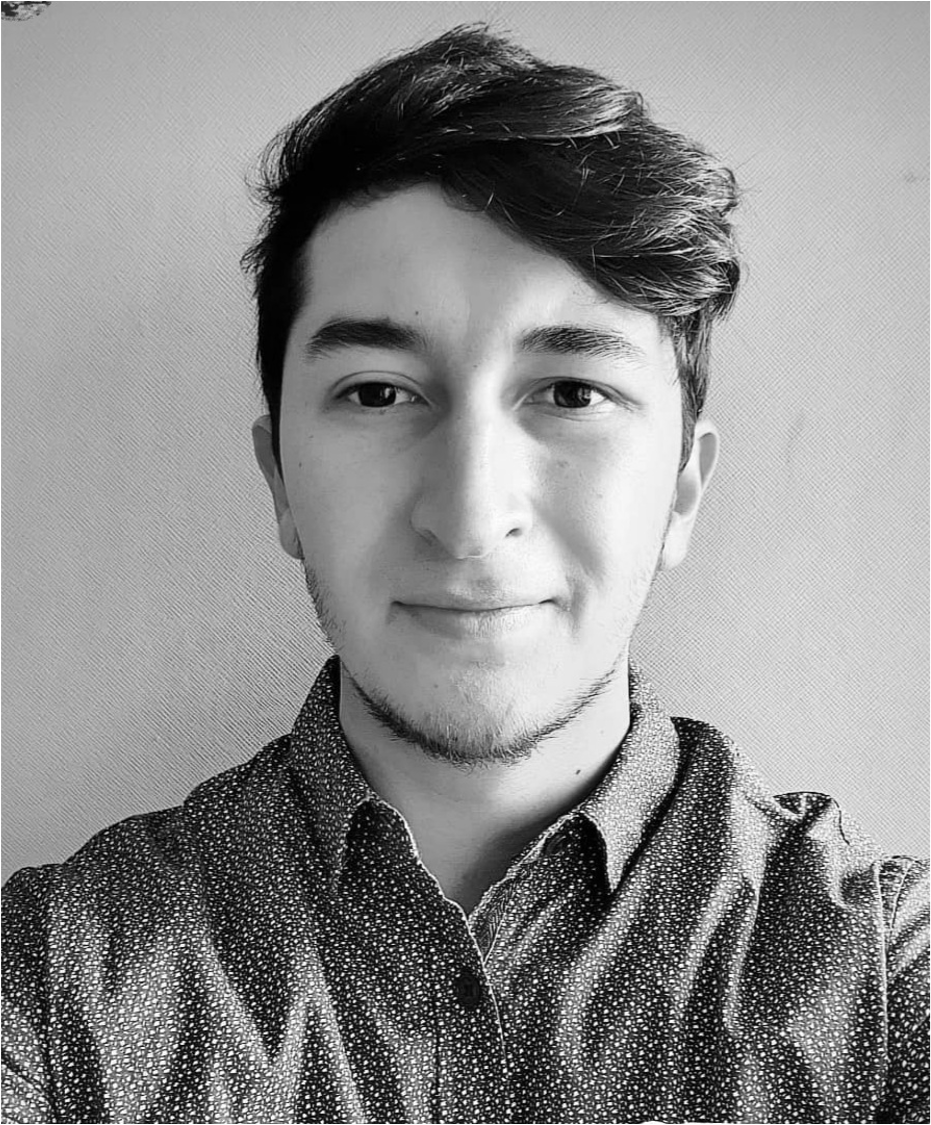}}]{Rodrigo Lara} received a B.S. in Computer Engineering from Universidad Andres Bello in 2019. Currently, he is a researcher at TOC Biometrics company. His main interests include computer vision, pattern recognition and deep learning applied to real problems such as synthetic images, super-resolution and segmentation.
\end{IEEEbiography}

\vspace{-0.3cm}
\begin{IEEEbiography}[{\includegraphics[width=1in,height=1.25in,clip,keepaspectratio]{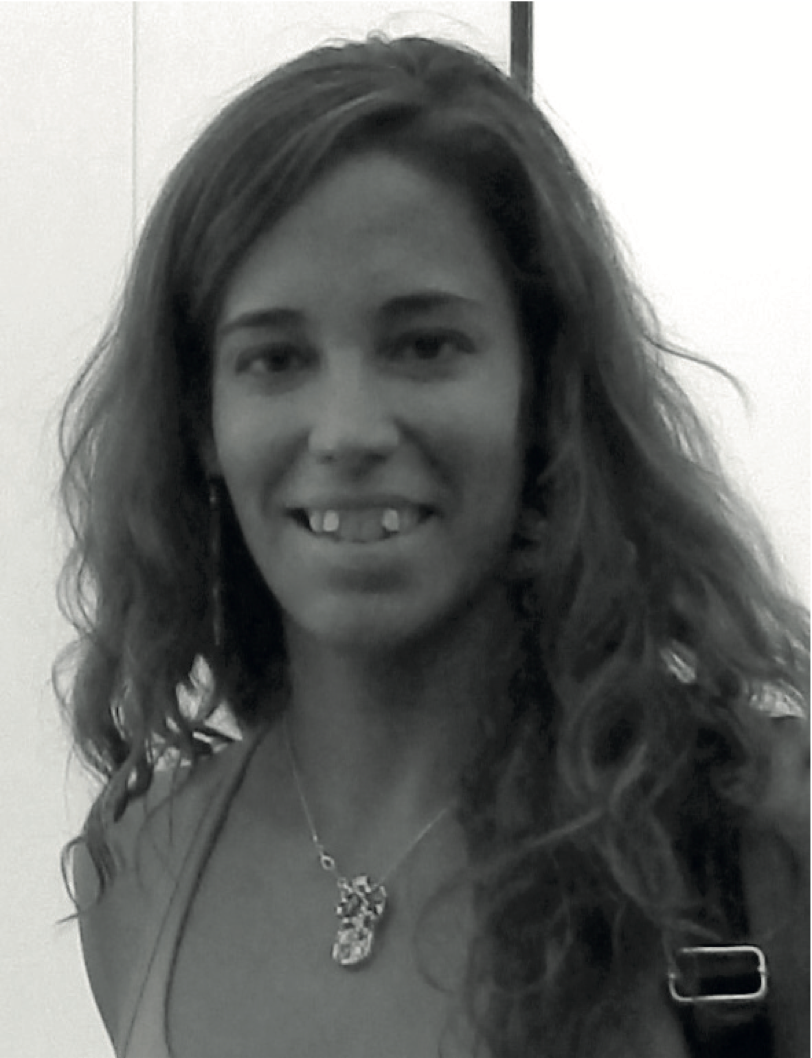}}]{Marta Gomez-Barrero} iis a Research Professor for IT-Security with a focus on biometric recognition at the Hochschule Ansbach, in Germany. Between 2016 and 2020, she was a postdoctoral researcher at the National Research Center for Applied Cybersecurity (ATHENE) - Hochschule Darmstadt, Germany. Before that, she received her MSc degrees in Computer Science and Mathematics (2011), and her PhD degree in Electrical Engineering (2016), all from Universidad Autonoma de Madrid, Spain. Her current research focuses on security and privacy evaluations of biometric systems, Presentation Attack Detection (PAD) methodologies, and biometric template protection (BTP) schemes. She has co-authored more than 90 publications, she is general chair of the BIOSIG conference, Co-Chair of the European Association for Biometrics Academic SIG, she is associate editor for the EURASIP Journal on Information Security, and represents the German Institute for Standardisation (DIN) in ISO/IEC SC37 JTC1 SC37 on biometrics. She has also received a number of distinctions, including: EAB European Biometric Industry Award 2015, Best Ph.D. Thesis Award by Universidad Autonoma de Madrid 2015/16, Archimedes Award for young researches from Spanish MECD, Best Paper Award at WIFS 2021, Siew-Sngiem Best Paper Award at ICB 2015, and Best Poster Award at ICB 2013.
\end{IEEEbiography}

\begin{IEEEbiography}[{\includegraphics[width=1in,height=1.25in,clip,keepaspectratio]{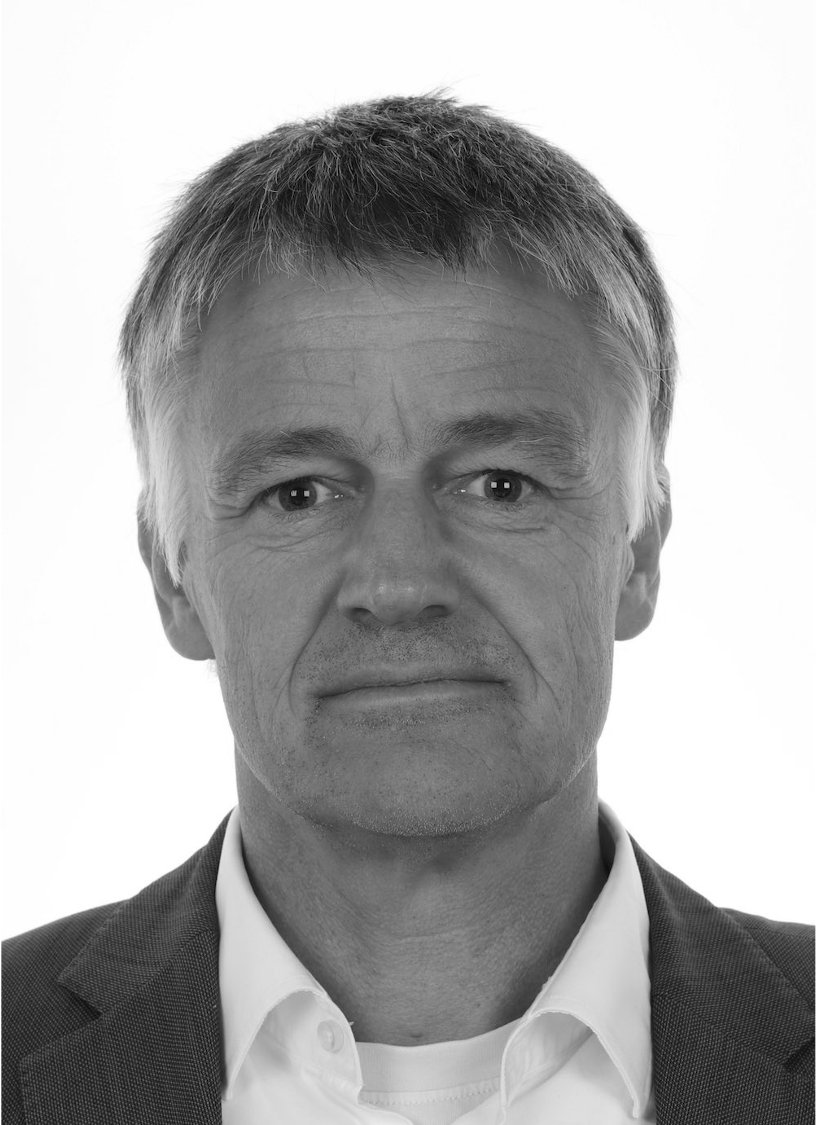}}]{Christoph Busch} is member of the Department of Information Security and Communication Technology (IIK) at the Norwegian University of Science and Technology (NTNU), Norway. He holds a joint appointment with the computer science faculty at Hochschule Darmstadt (HDA), Germany. Further he lectures the course Biometric Systems at Denmark’s DTU since 2007. On behalf of the German BSI he has been the coordinator for the project series BioIS, BioFace, BioFinger, BioKeyS Pilot-DB, KBEinweg and NFIQ2.0. In the European research program he was initiator of the Integrated Project 3D-Face, FIDELITY and iMARS. Further he was/is partner in the projects TURBINE, BEST Network, ORIGINS, INGRESS, PIDaaS, SOTAMD, RESPECT and TReSPAsS. He is also principal investigator in the German National Research Center for Applied Cybersecurity (ATHENE). Moreover Christoph Busch is co-founder and member of board of the European Association for Biometrics (www.eab.org) that was established in 2011 and assembles in the meantime more than 200 institutional members. Christoph co-authored more than 600 technical papers and has been a speaker at international conferences. He is member of the editorial board of the IET journal.
\end{IEEEbiography}

\clearpage

\appendix

\section{Appendix A}
\label{sec:append}

\begin{figure}[h]
\centering 
\includegraphics[width=1.0\linewidth]{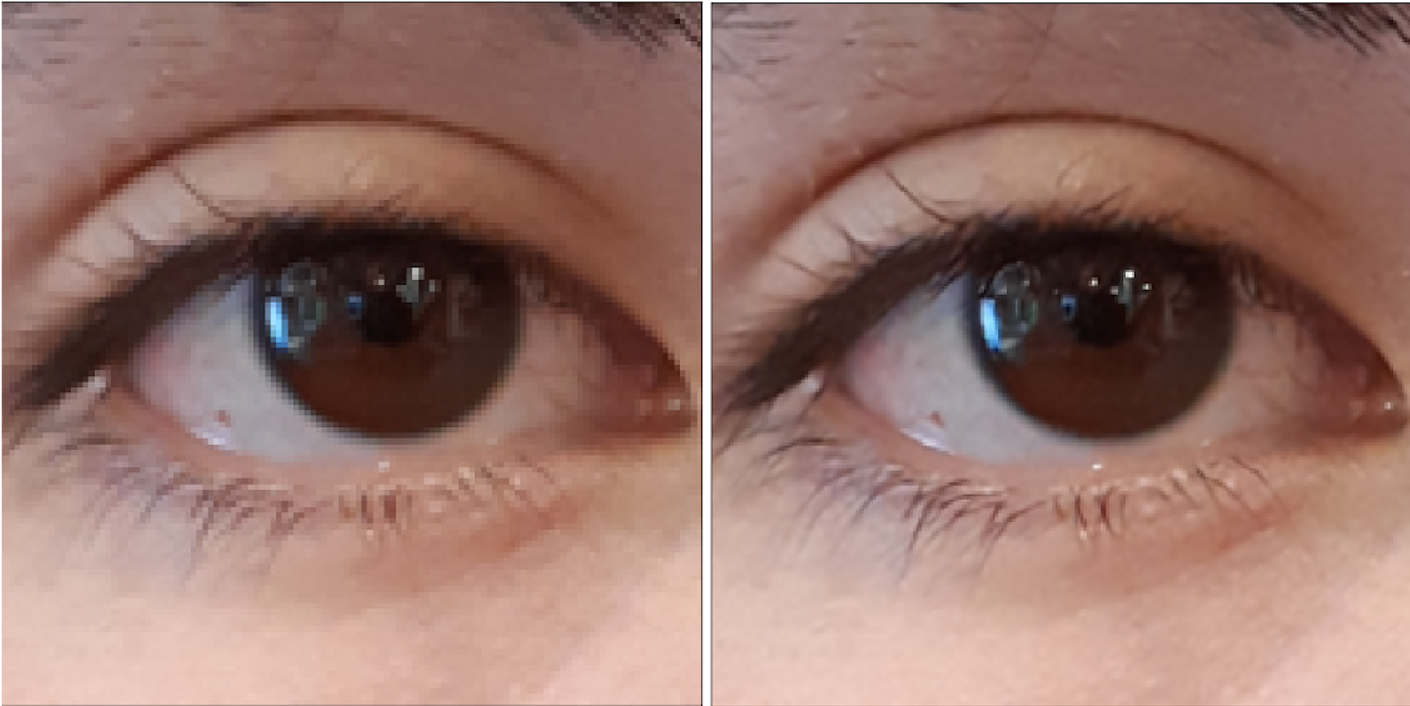}\\
\includegraphics[width=1.0\linewidth]{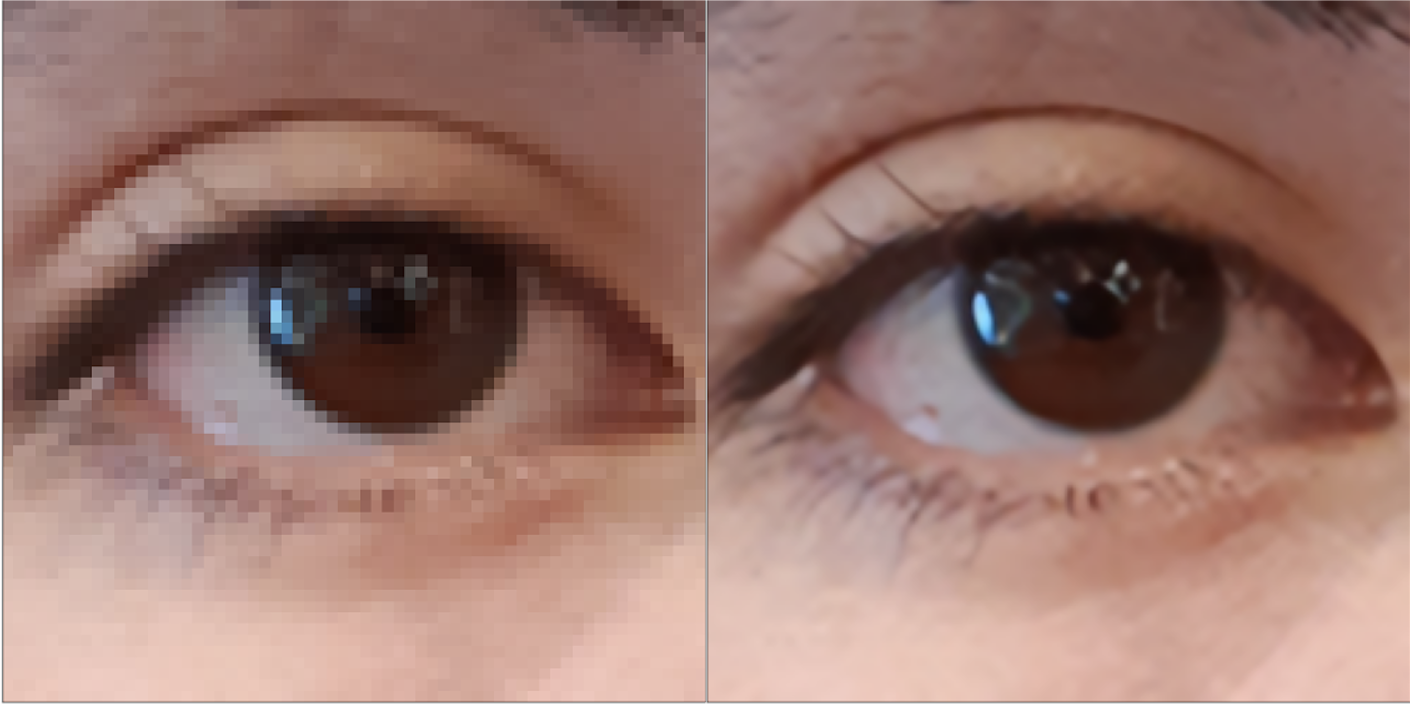}\\
\includegraphics[width=1.0\linewidth]{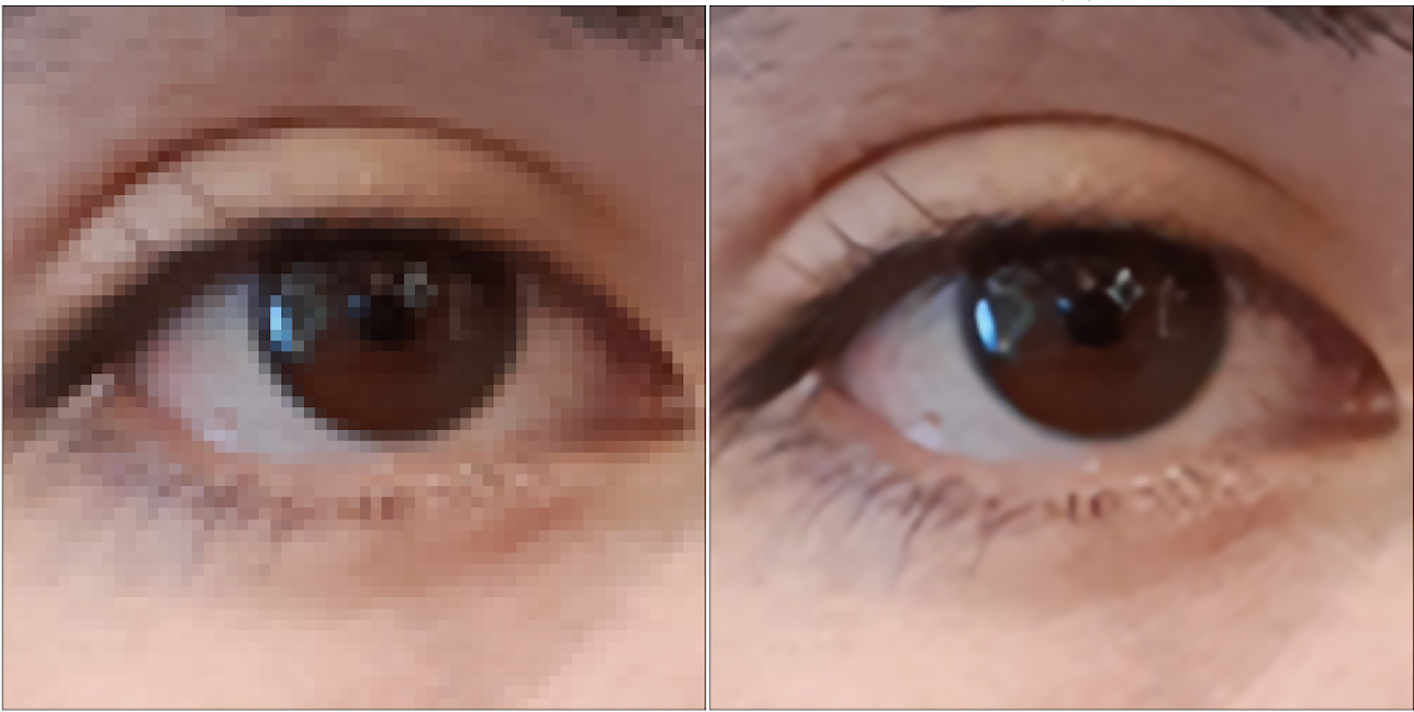}
\caption{Example of MObBIO Super resolution images. Top: x2. Middle: x3 and Bottom: x4. Increase the size of the images to see the effect of SR. } 
\label{fig:sr_example}
\end{figure}

Fig. \ref{fig:sr_example} show examples of the proposed SR method ESISR with super-resolution x2, x3 and x4, respectively.

\begin{figure}[t]
\centering 
\includegraphics[width=0.9\linewidth]{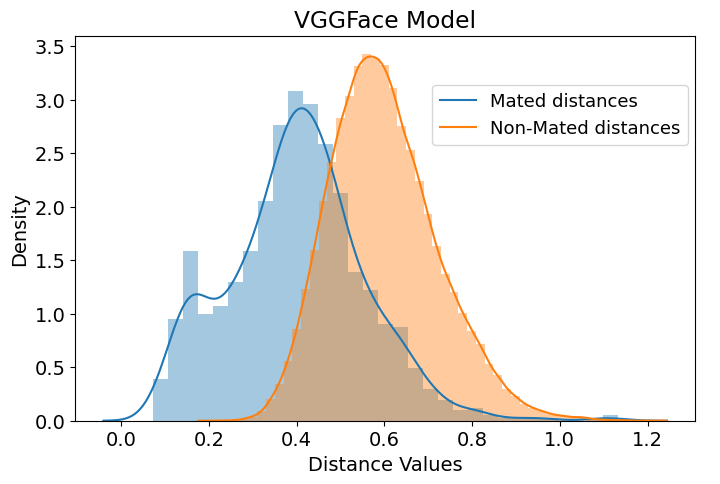}
\includegraphics[width=0.9\linewidth]{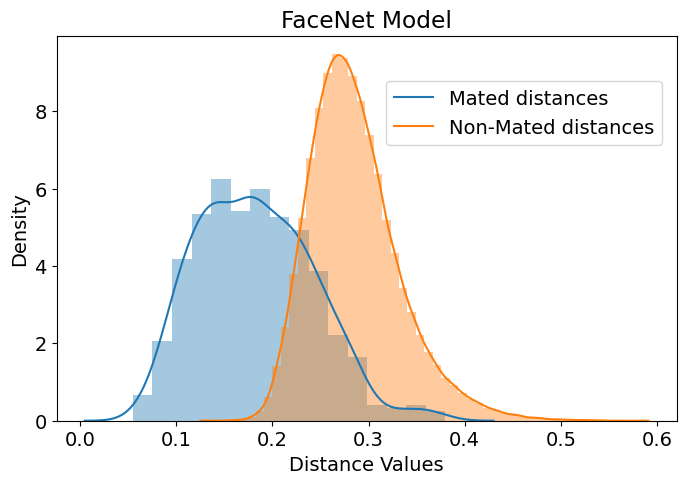}
\includegraphics[width=0.9\linewidth]{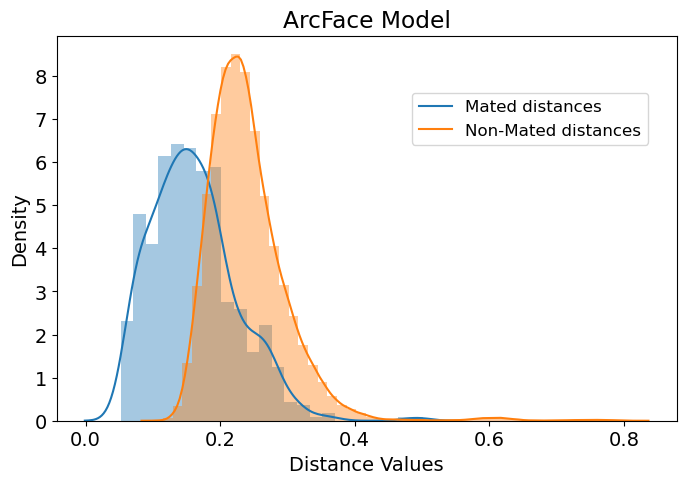}
\caption{Mated and Non-mated score distributions for FaceNet (top), VGGFace (middle) and ArcFace (bottom).} 
\centering
\label{distribution_new}
\end{figure}

Fig.~\ref{distribution_new} shows the probability density functions of the comparisons between mated and non-mated features vectors for the FaceNet, VGGFace, and ArcFace models. The VGGFace feature-vector is more spread between 0.1 and 1.0 in contrast to the FaceNet and ArcFace vectors, which are concentrated between 0.1 and 0.4. All distributions shown some overlap, which in turn leads to the non-perfect verification rates presented in the following.

\begin{figure*}[]
\centering 
\includegraphics[width=0.32\linewidth]{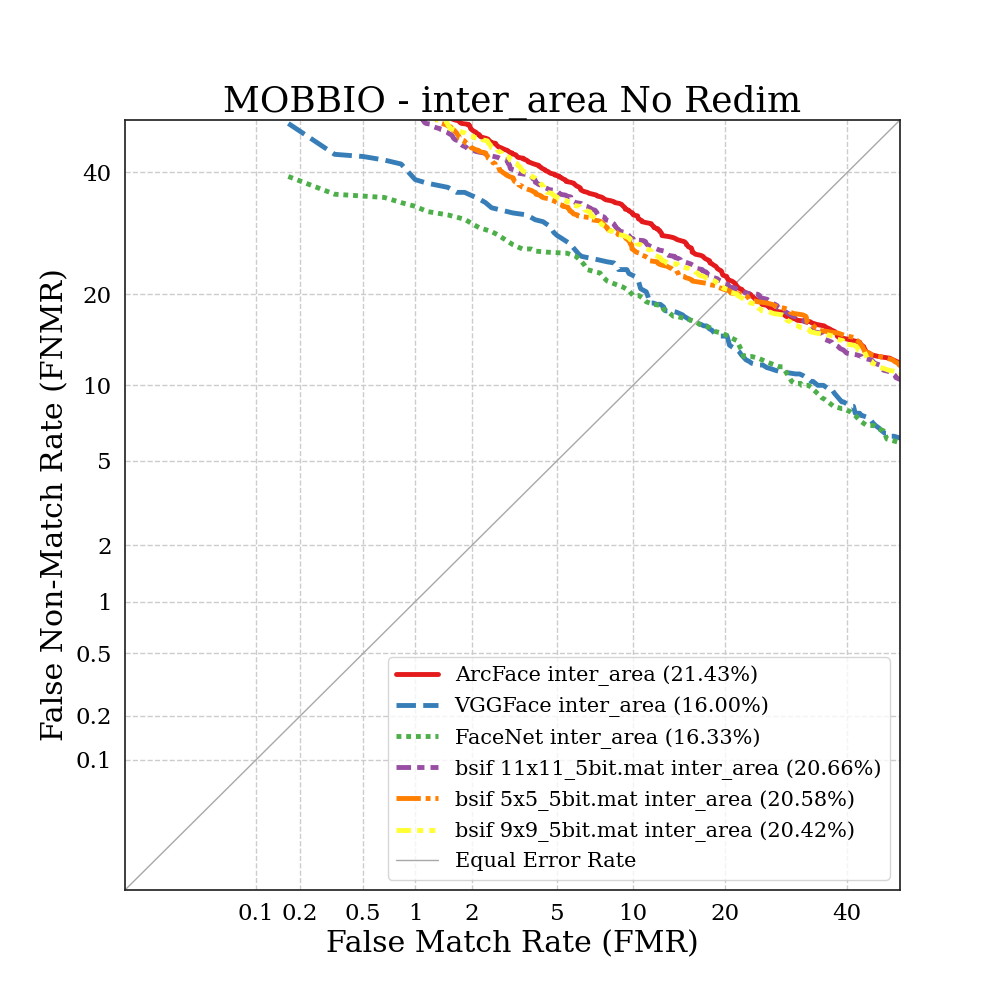}
\includegraphics[width=0.32\linewidth]{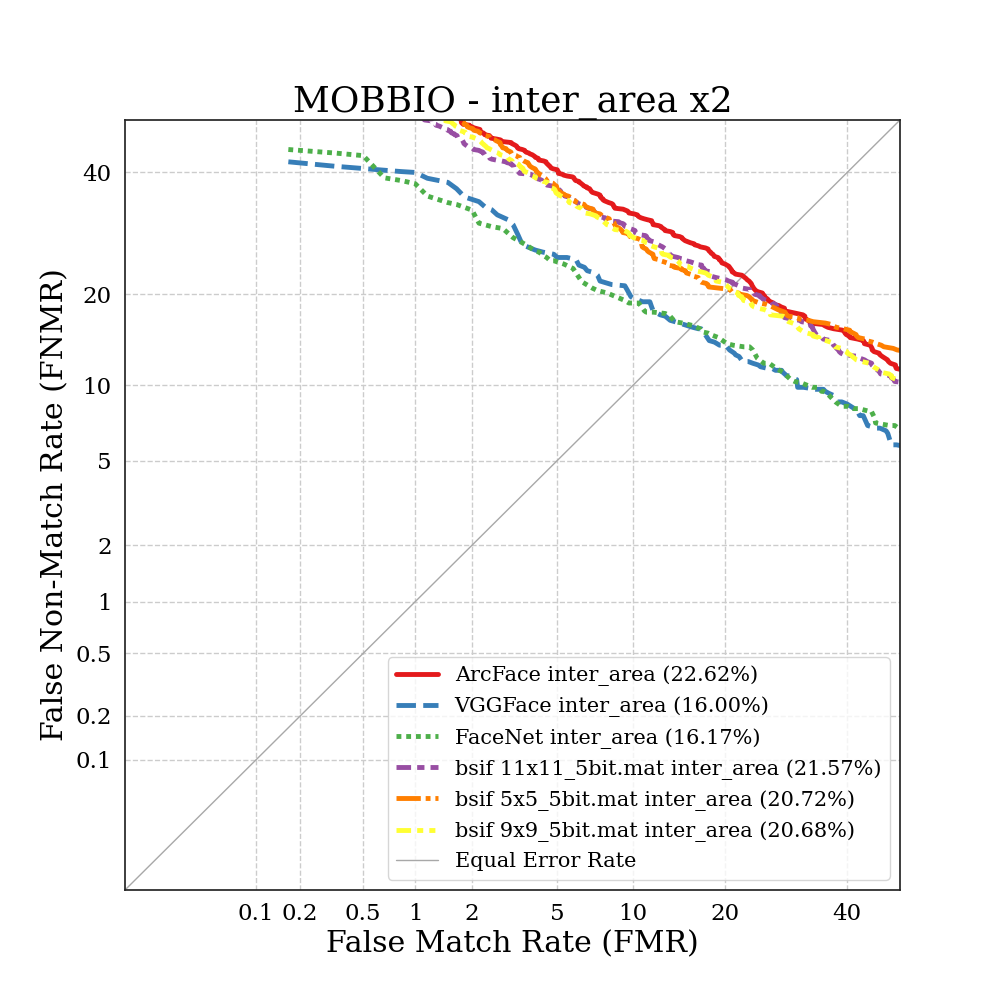}
\includegraphics[width=0.32\linewidth]{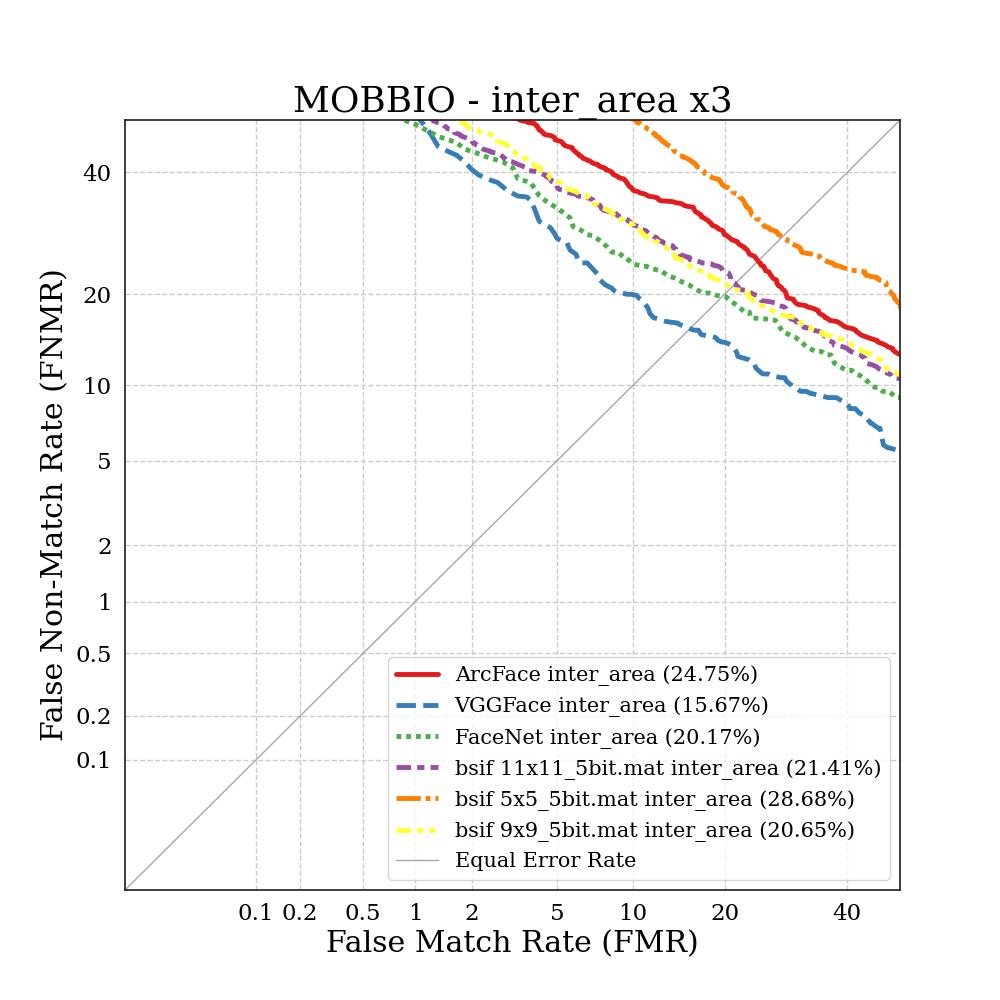}
\includegraphics[width=0.32\linewidth]{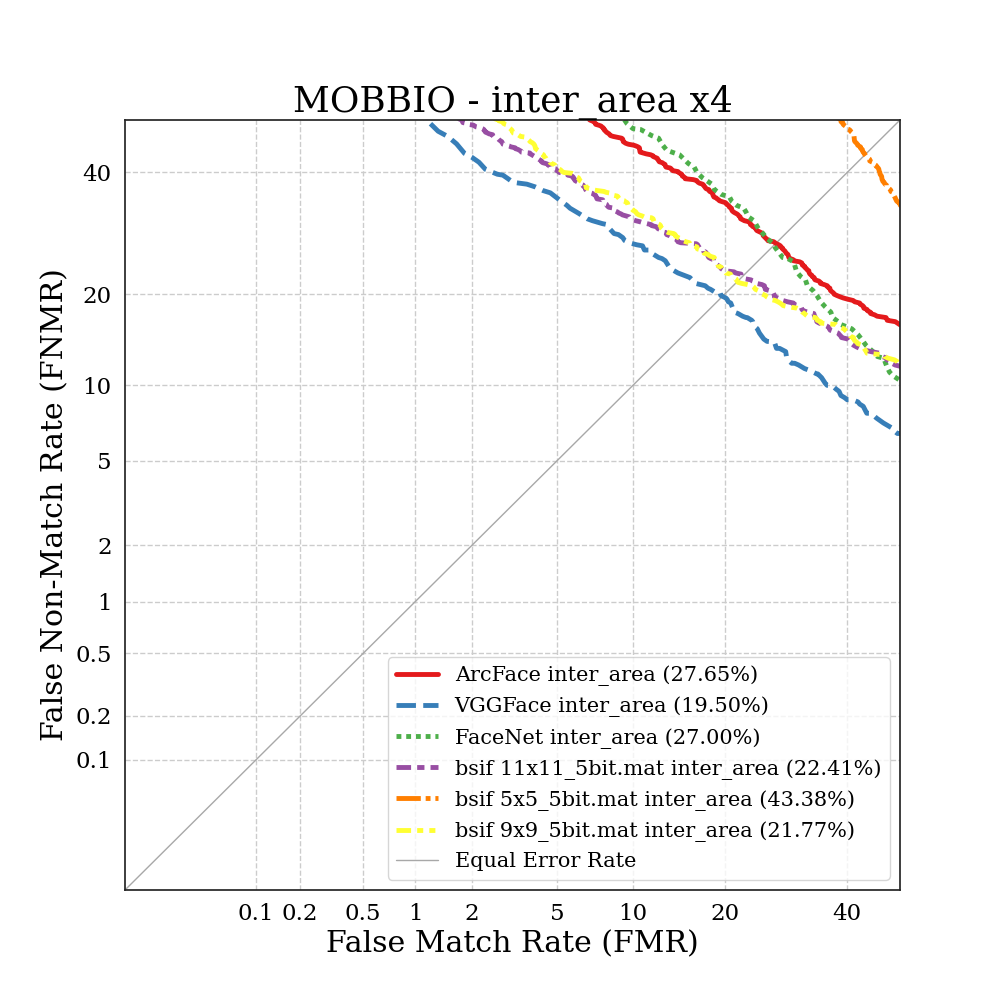}
\includegraphics[width=0.32\linewidth]{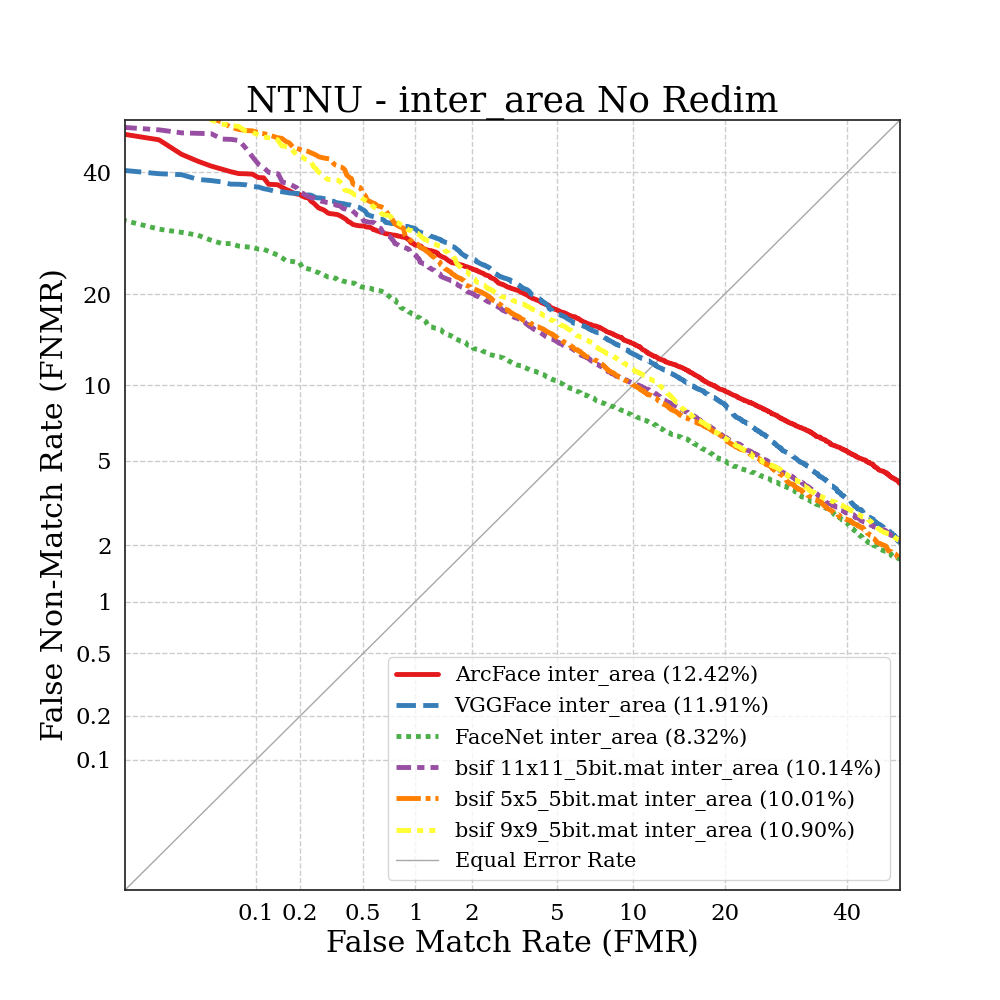}
\includegraphics[width=0.32\linewidth]{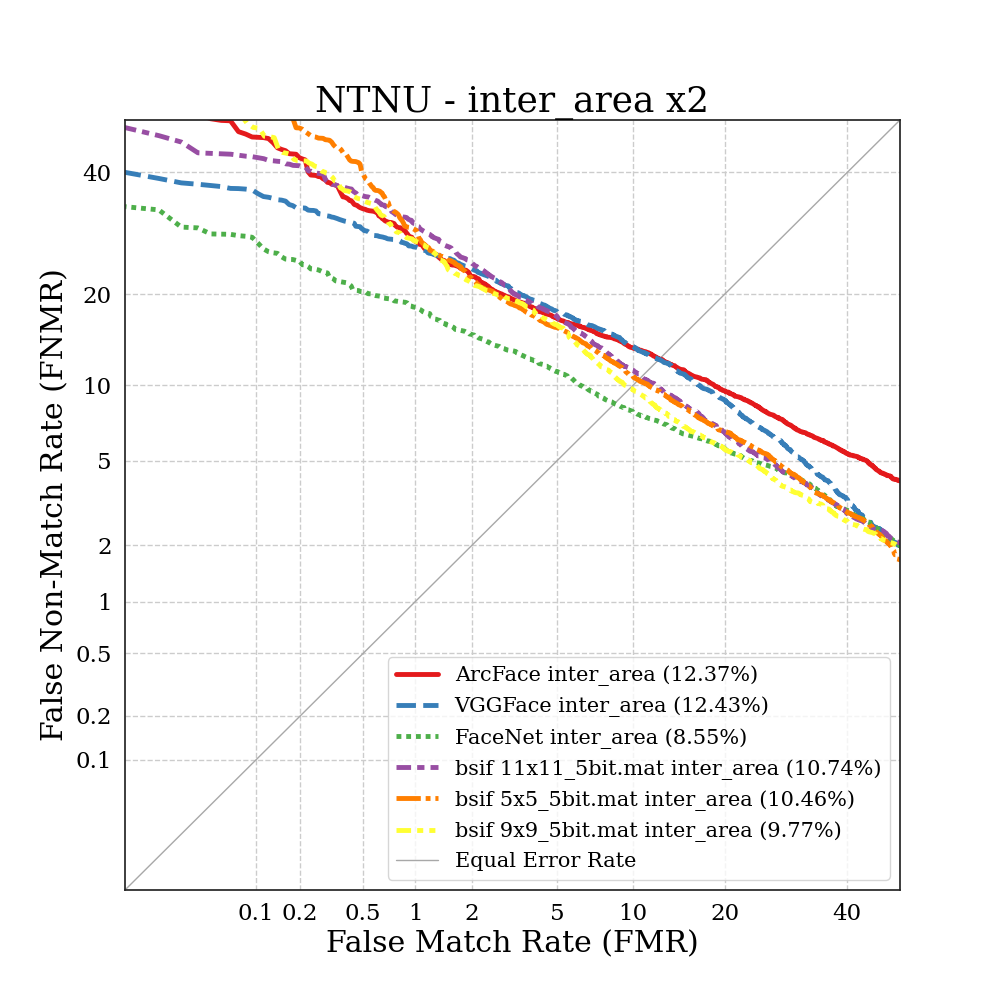}
\includegraphics[width=0.32\linewidth]{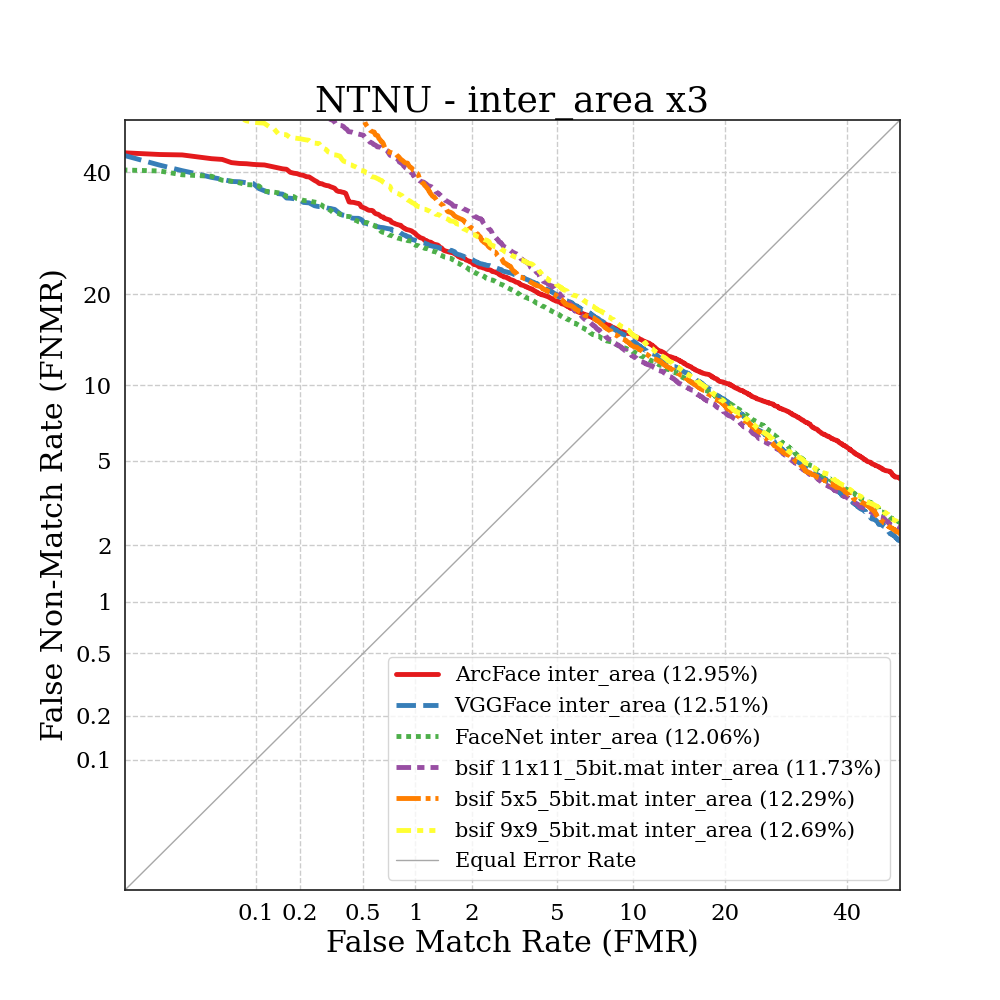}
\includegraphics[width=0.32\linewidth]{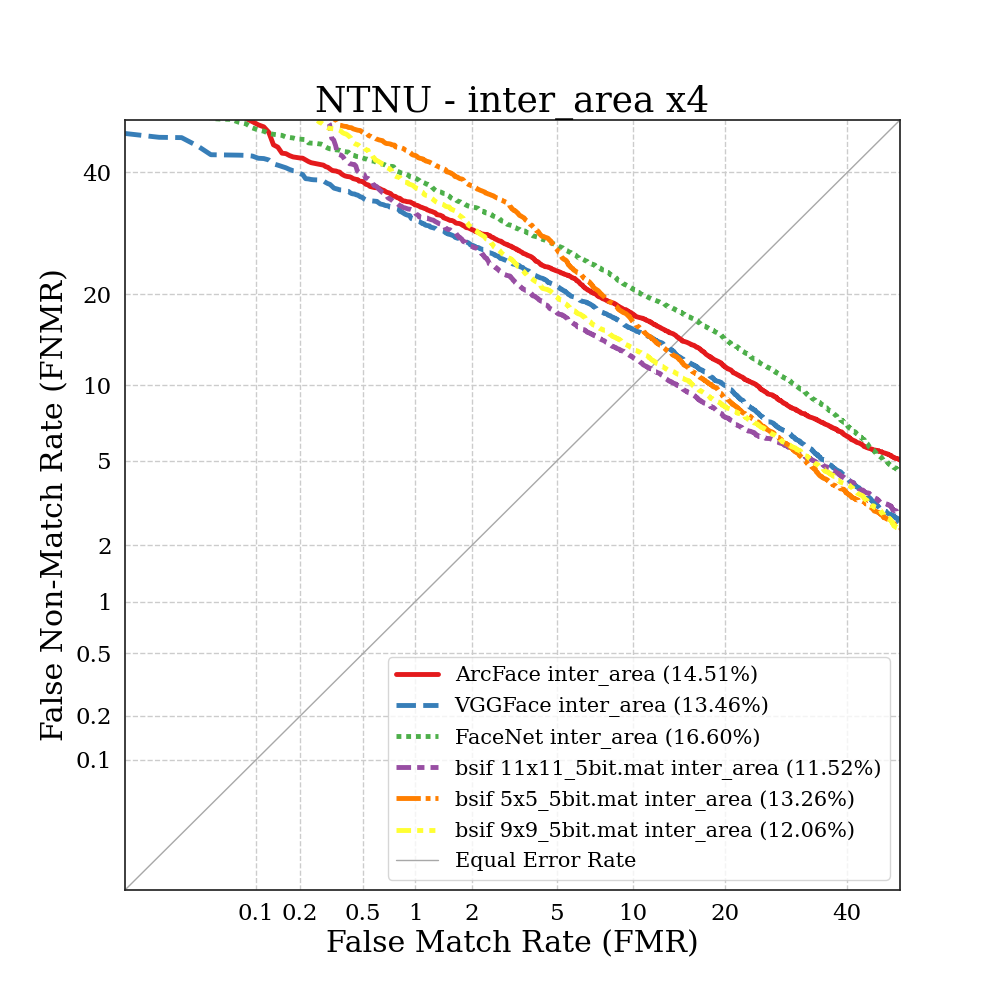}
\caption{DETs for MObBIO and NTNU dataset including selfie recognition systems based on traditional Inter-Area resizing. The EER is showed in parenthesis for each method}. 
\label{DETs-mobio-ntu-area}
\end{figure*}

\begin{figure*}[]
\centering 
\includegraphics[width=0.32\linewidth]{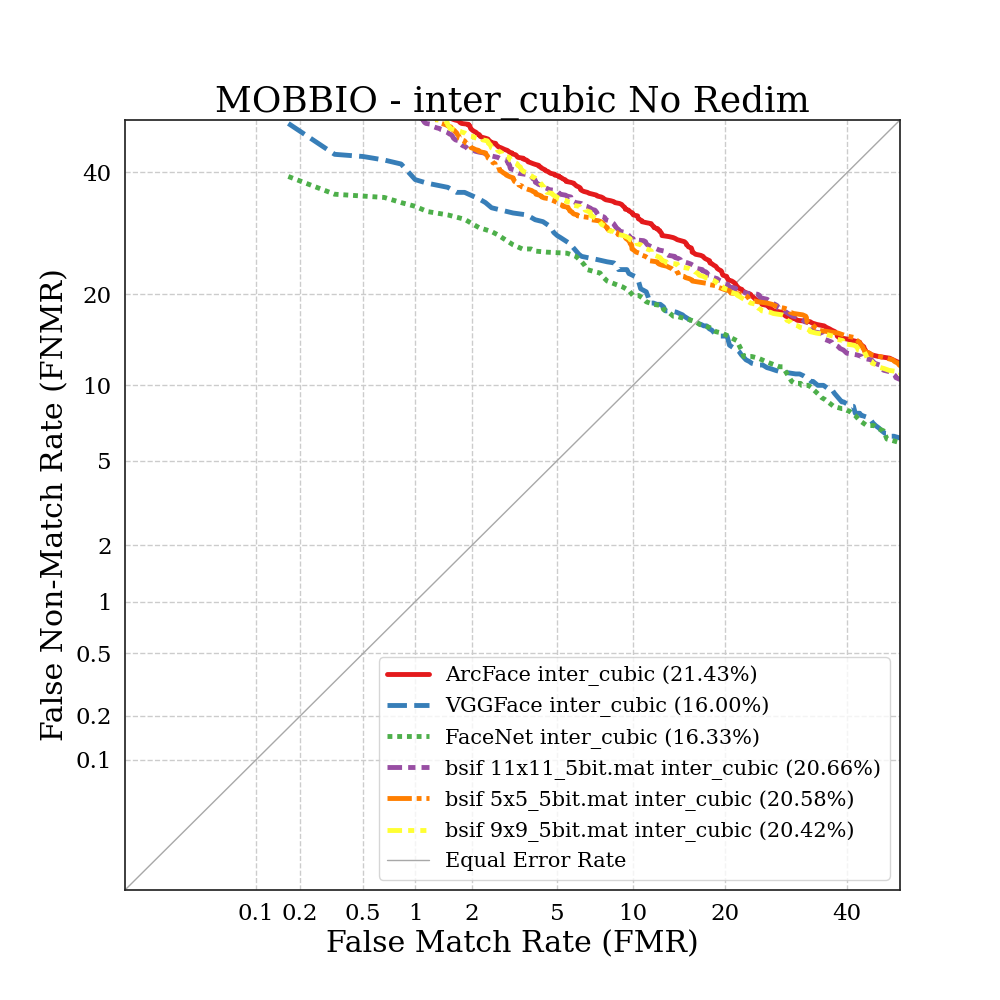}
\includegraphics[width=0.32\linewidth]{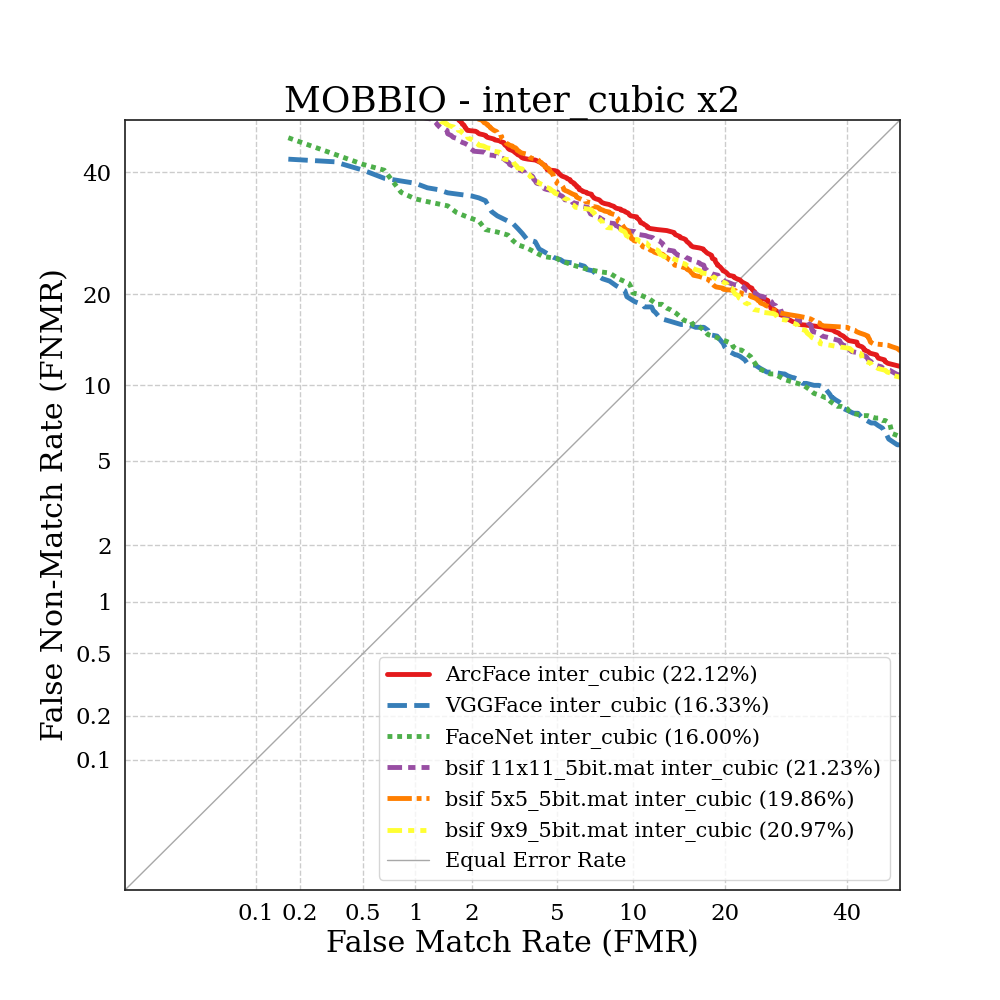}
\includegraphics[width=0.32\linewidth]{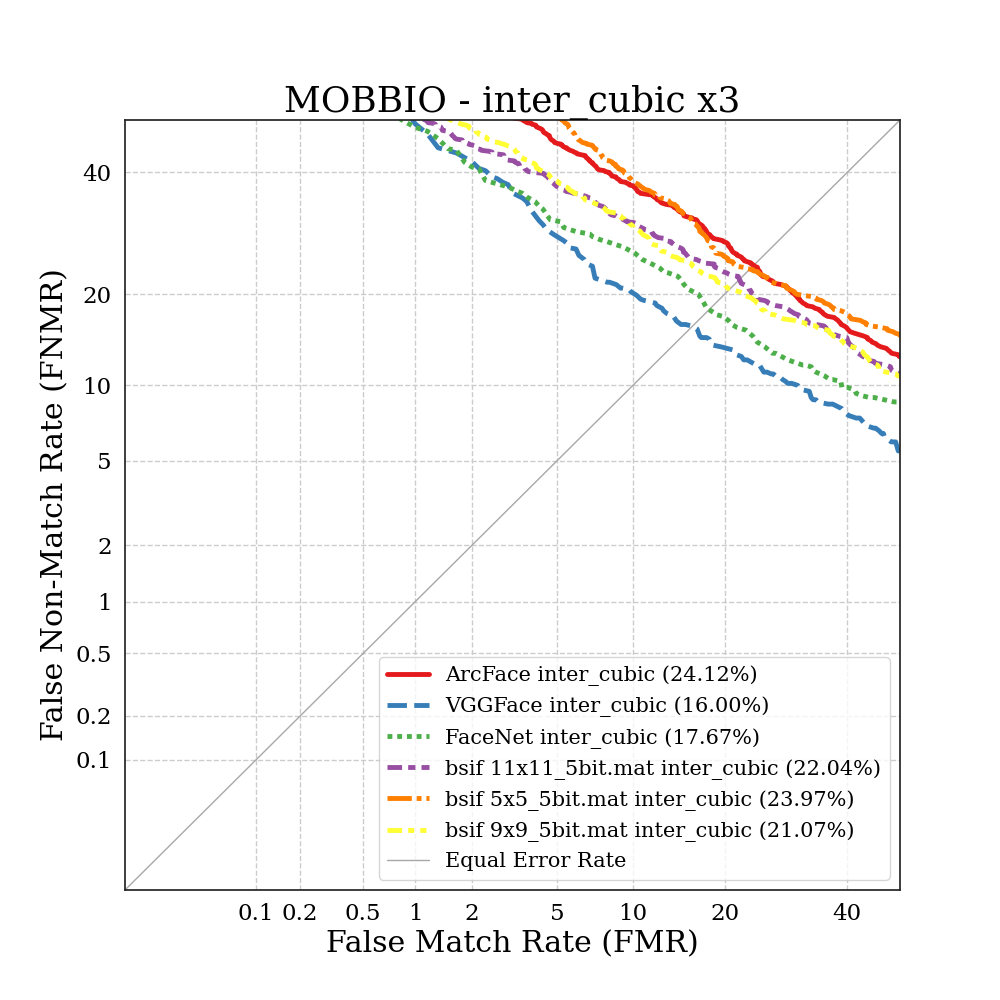}
\includegraphics[width=0.32\linewidth]{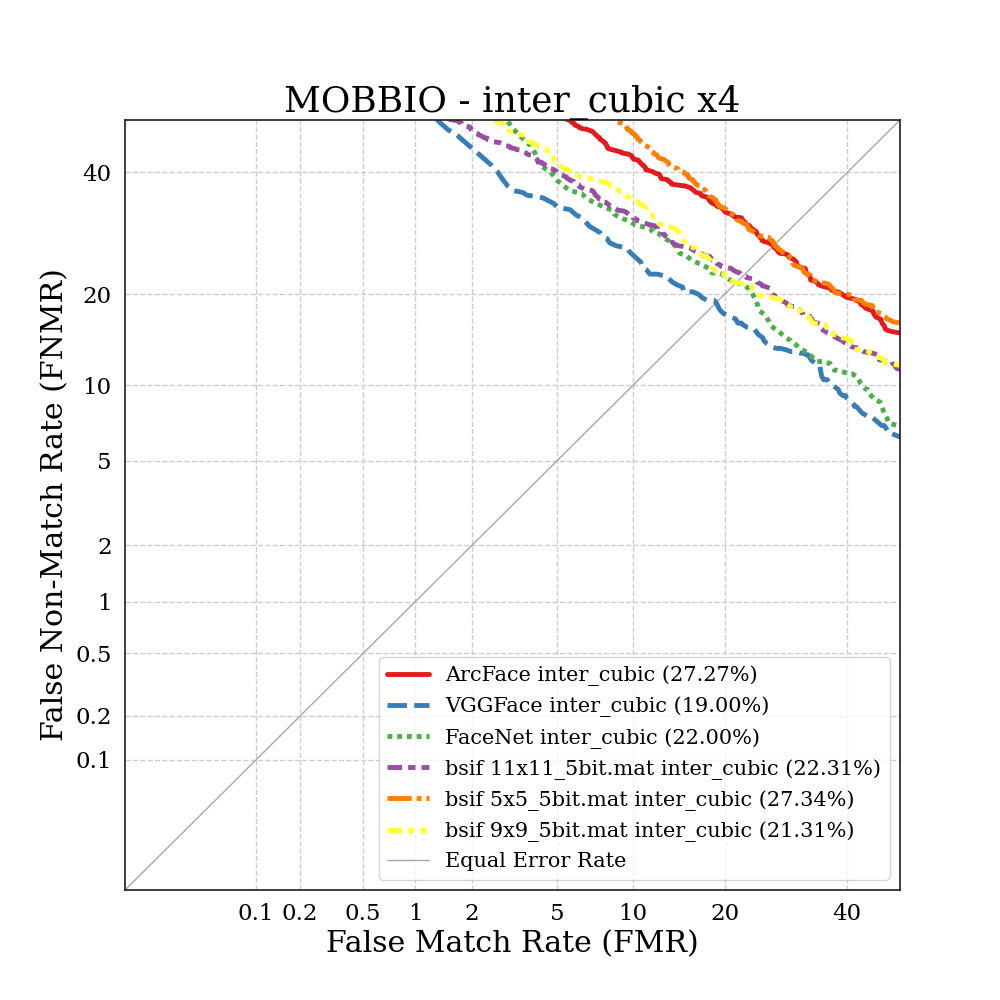}
\includegraphics[width=0.32\linewidth]{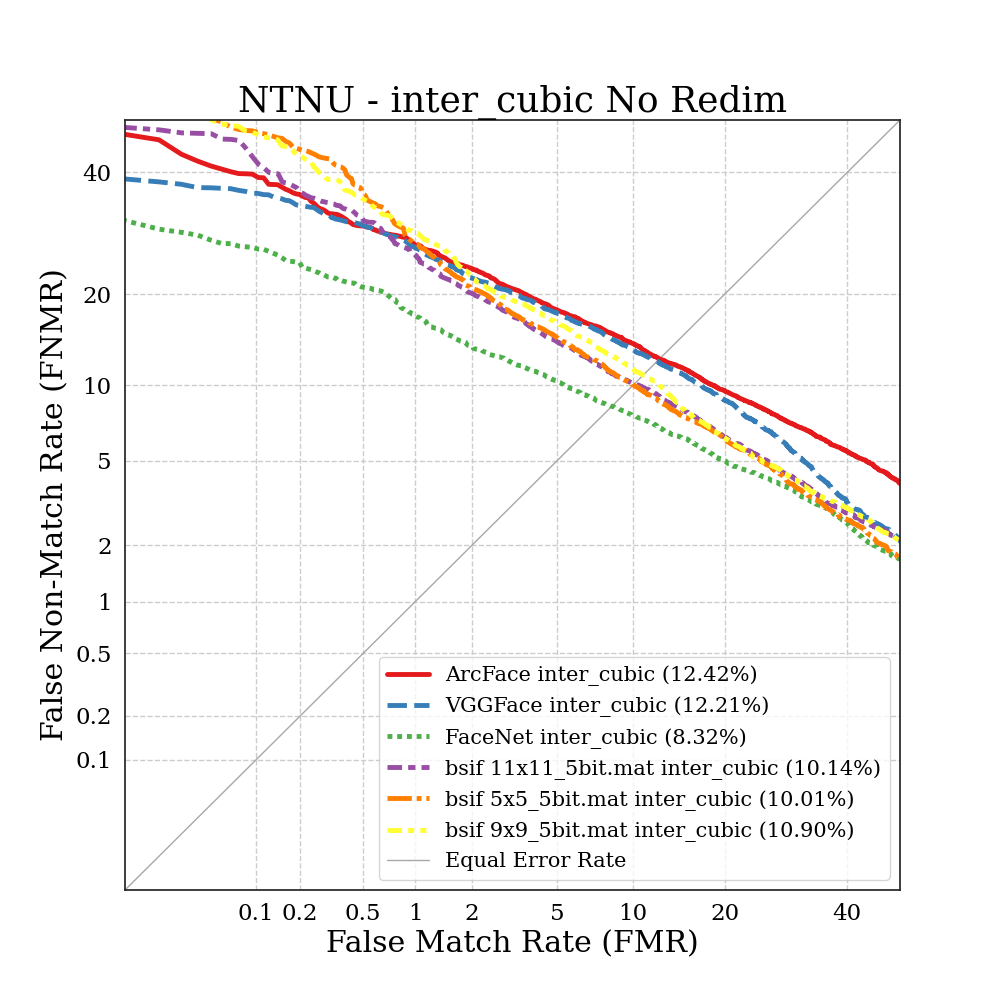}
\includegraphics[width=0.32\linewidth]{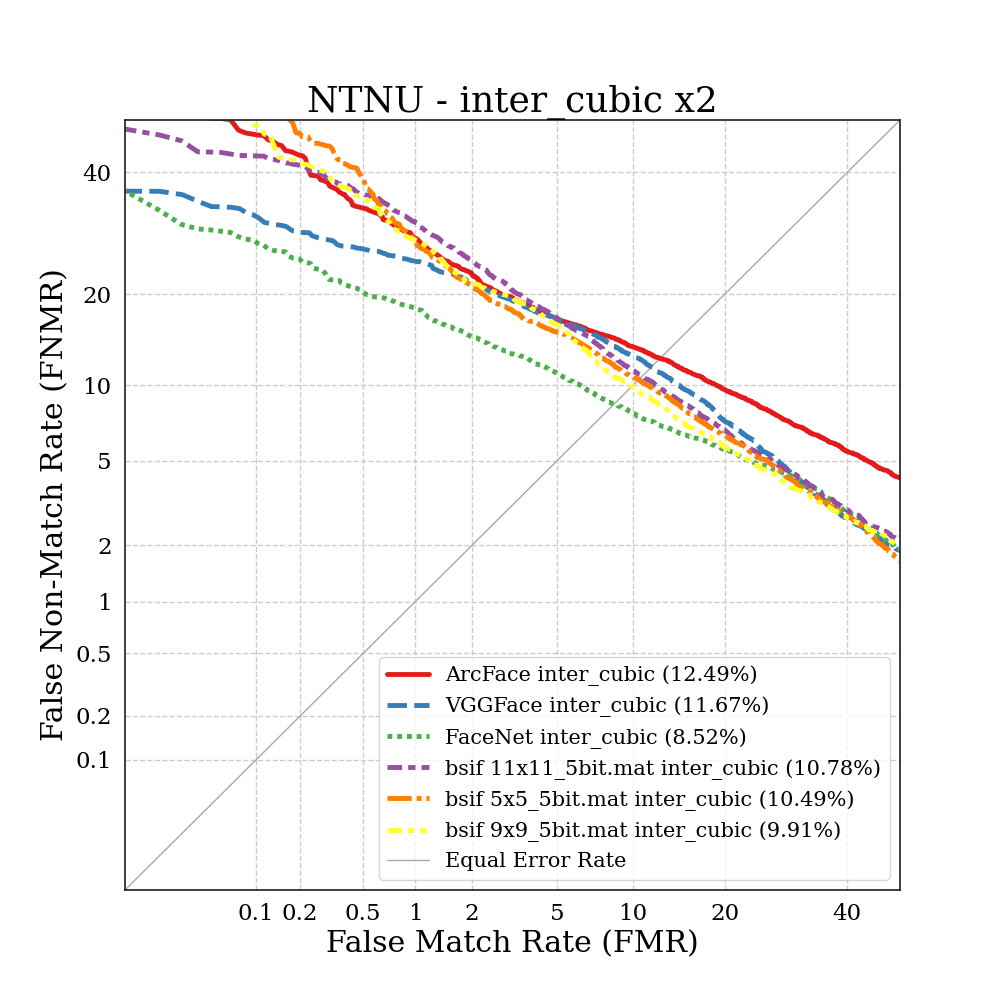}
\includegraphics[width=0.32\linewidth]{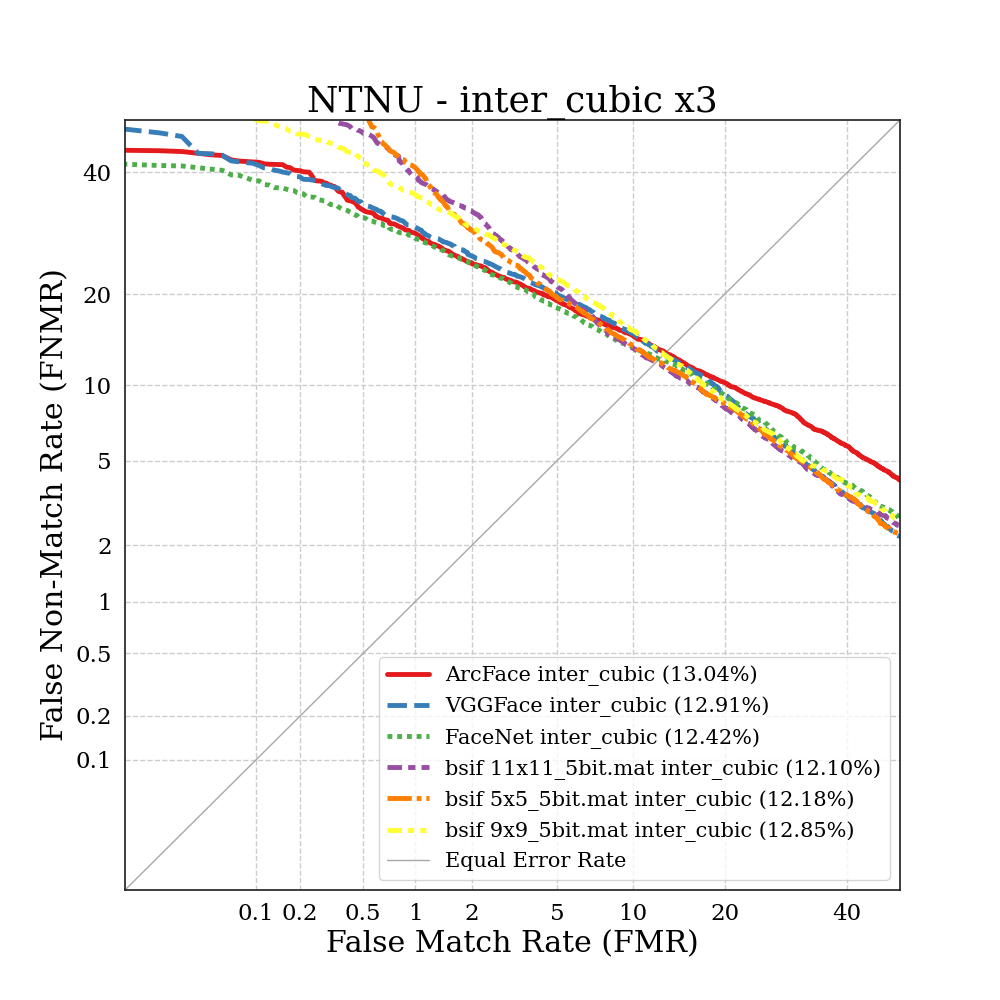}
\includegraphics[width=0.32\linewidth]{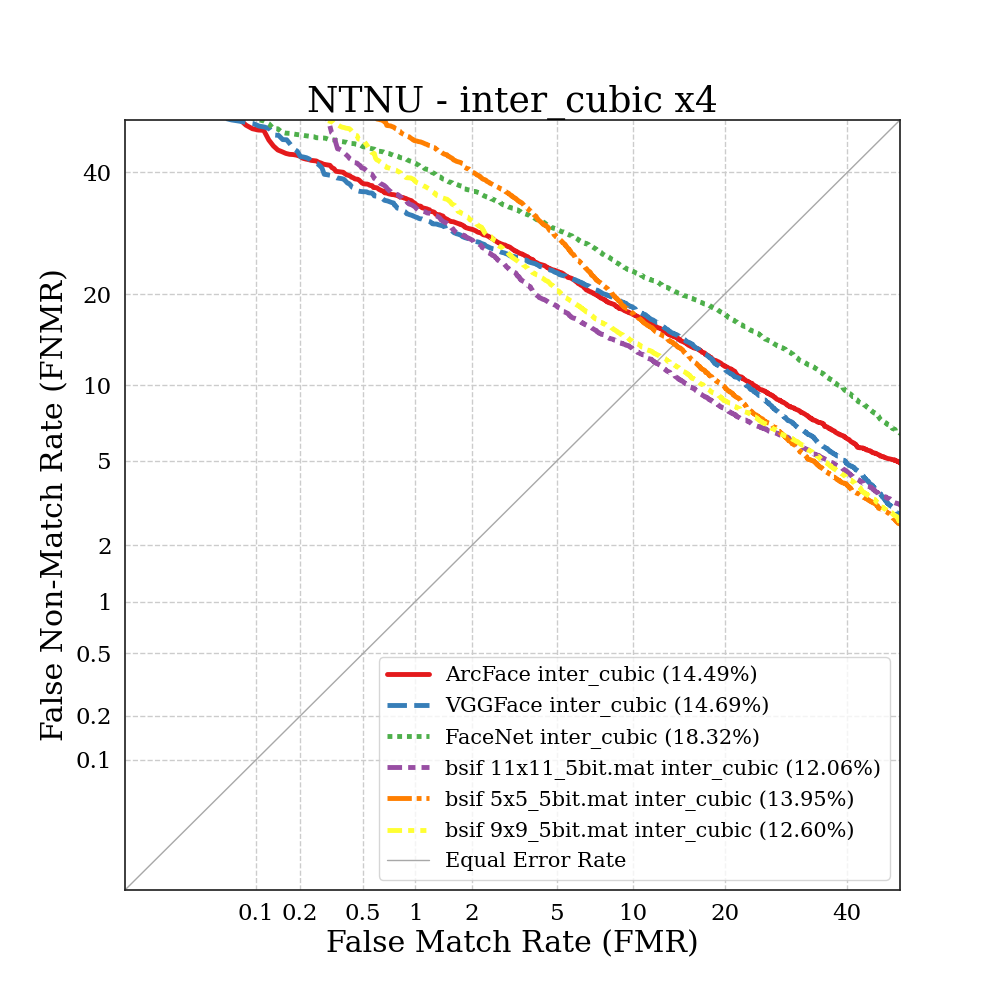}
\caption{DETs for MObBIO and NTNU dataset including selfie recognition systems based on traditional Inter-Cubic resizing. The EER is showed in parenthesis for each method}. 
\label{DETs-mobio-ntu-cubic}
\end{figure*}

\begin{figure*}[]
\centering 
\includegraphics[width=0.32\linewidth]{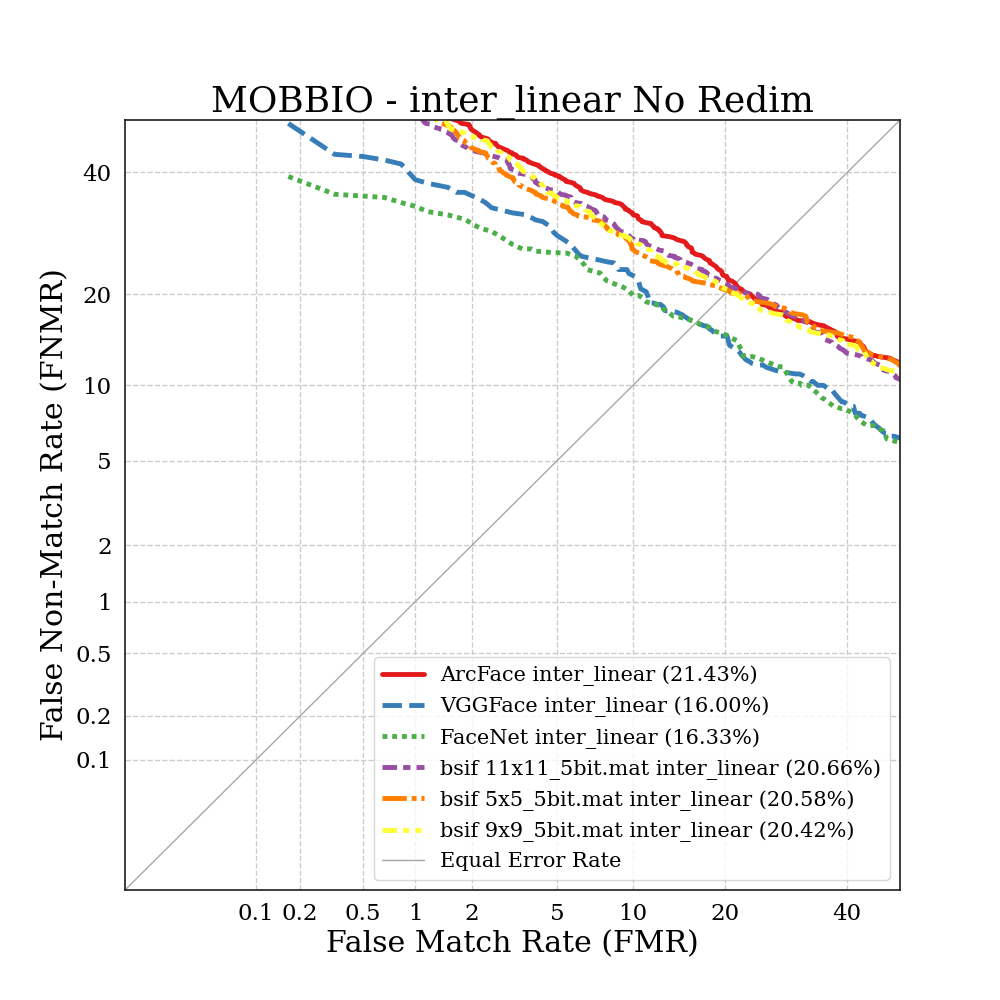}
\includegraphics[width=0.32\linewidth]{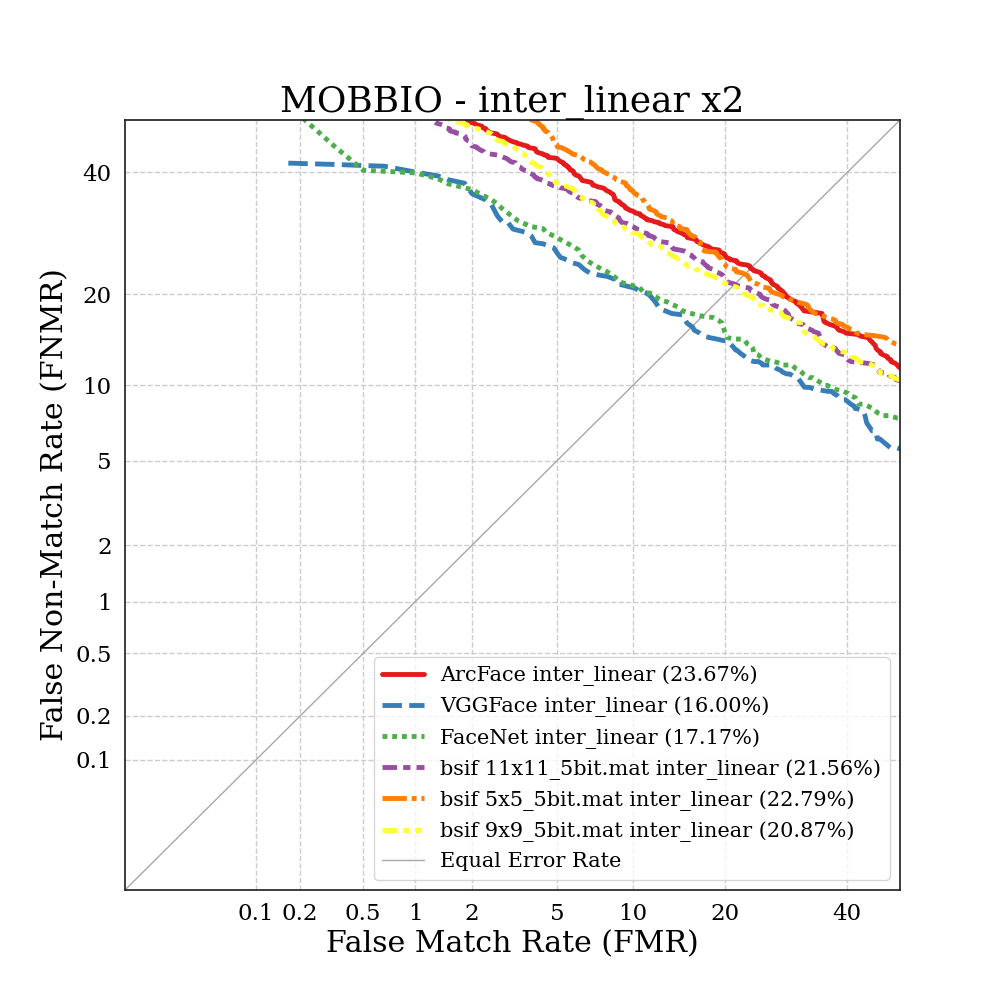}
\includegraphics[width=0.32\linewidth]{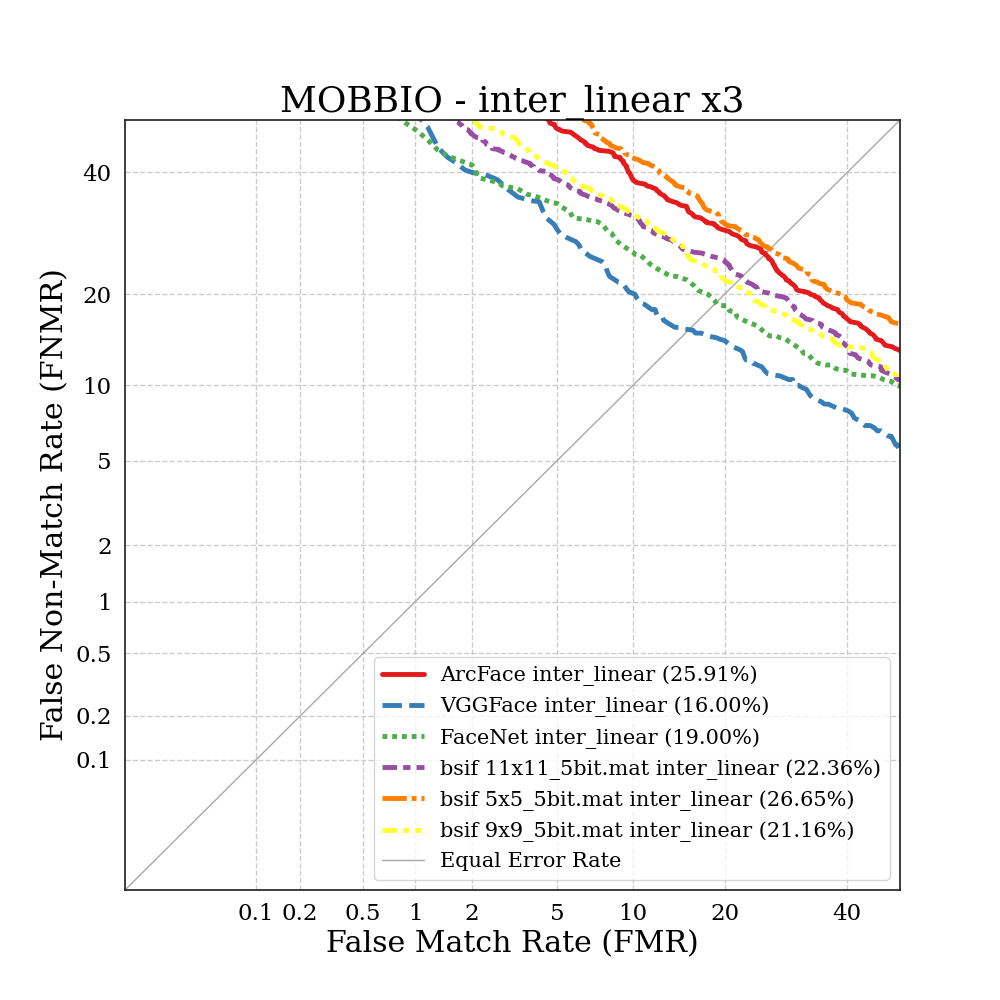}
\includegraphics[width=0.32\linewidth]{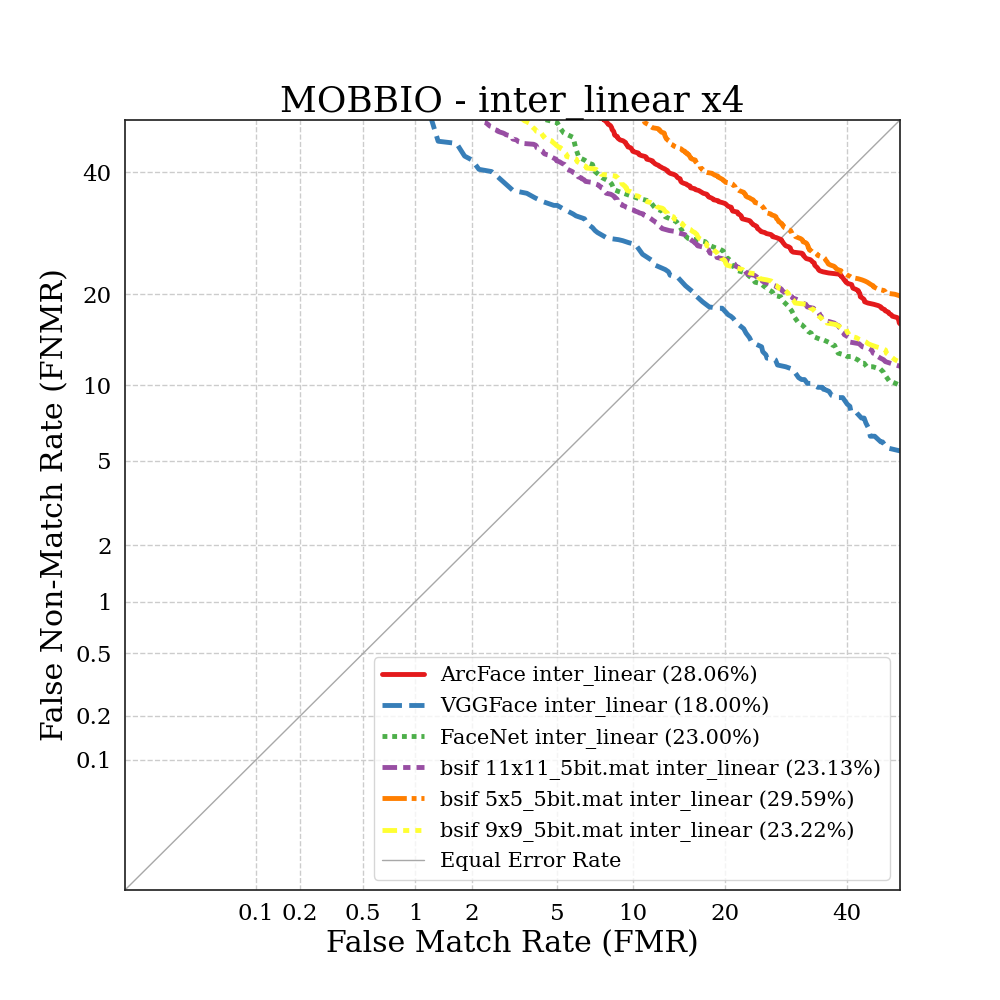}
\includegraphics[width=0.32\linewidth]{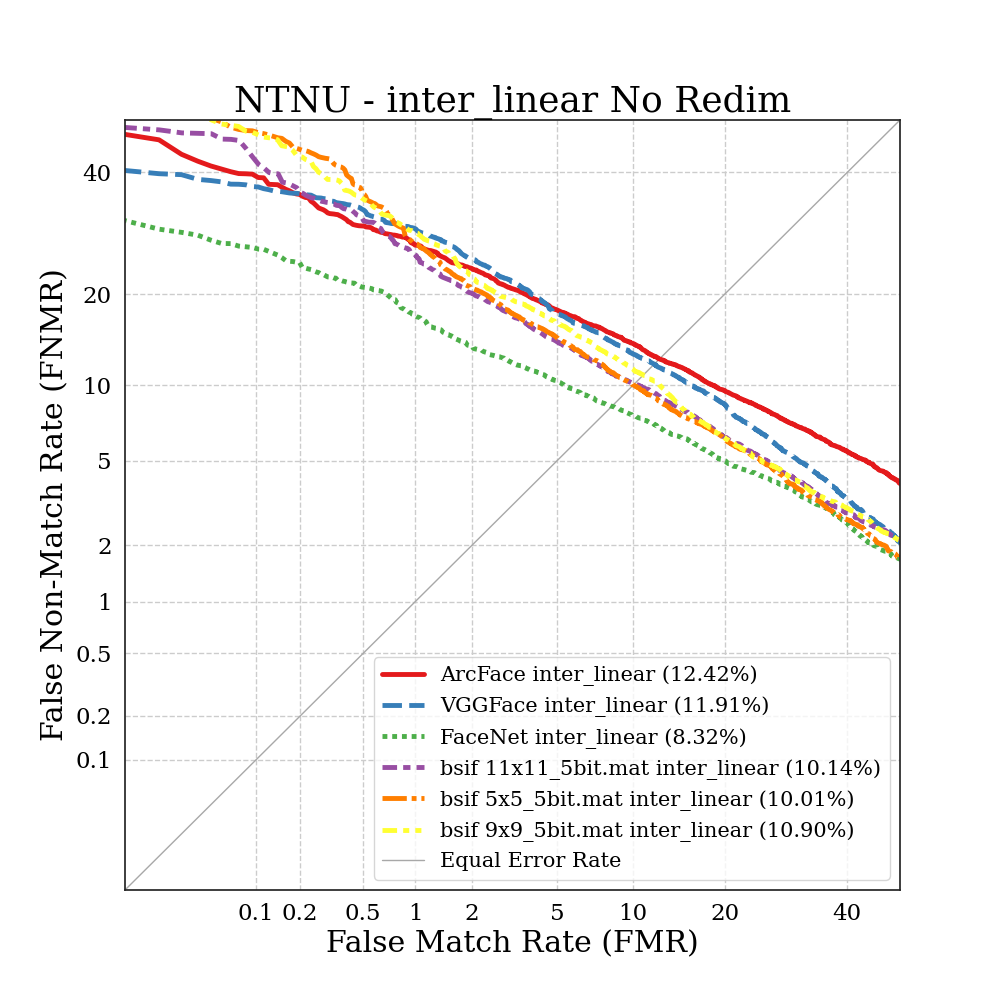}
\includegraphics[width=0.32\linewidth]{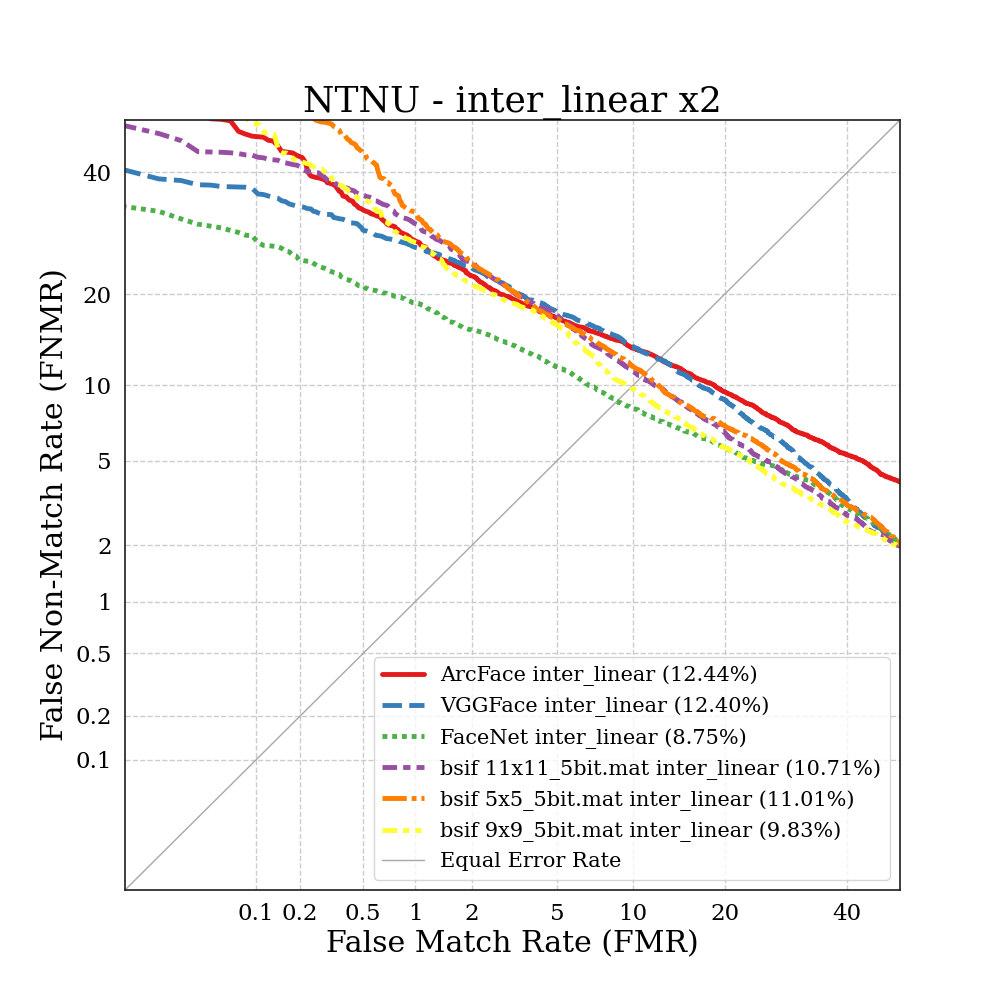}
\includegraphics[width=0.32\linewidth]{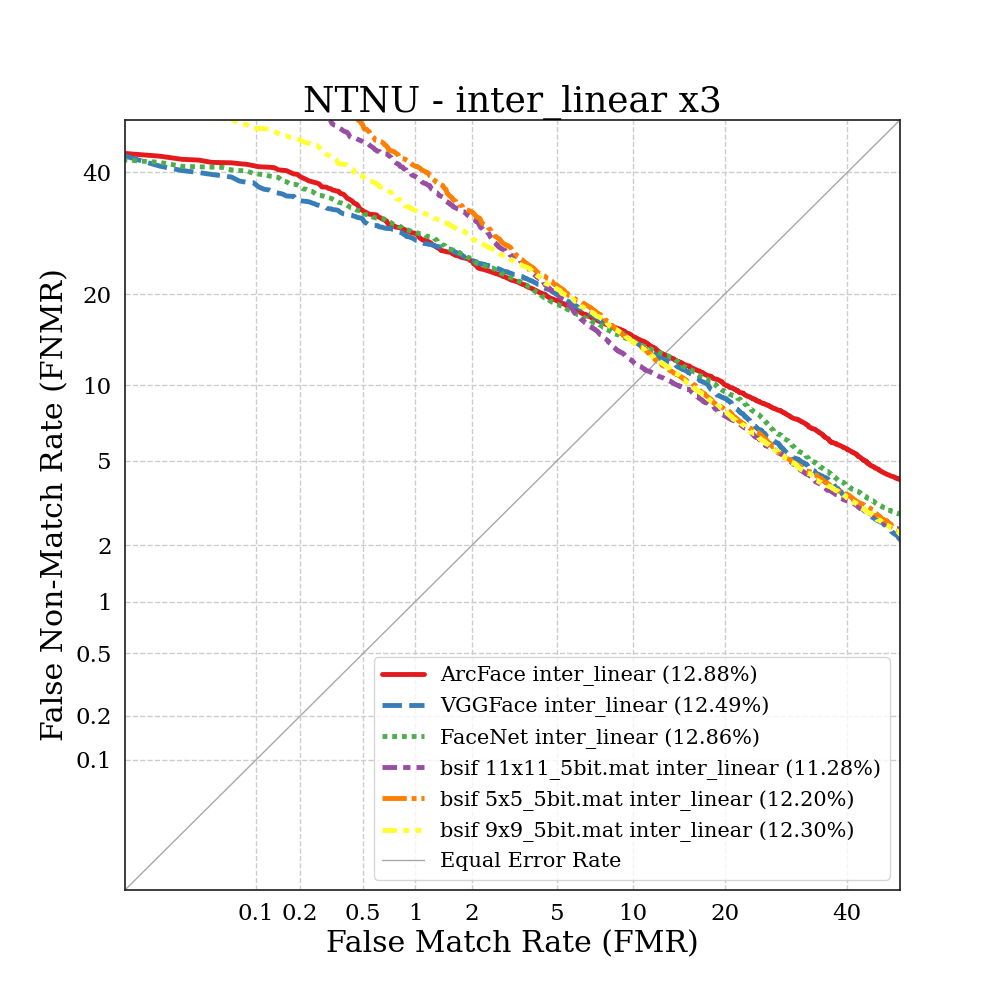}
\includegraphics[width=0.32\linewidth]{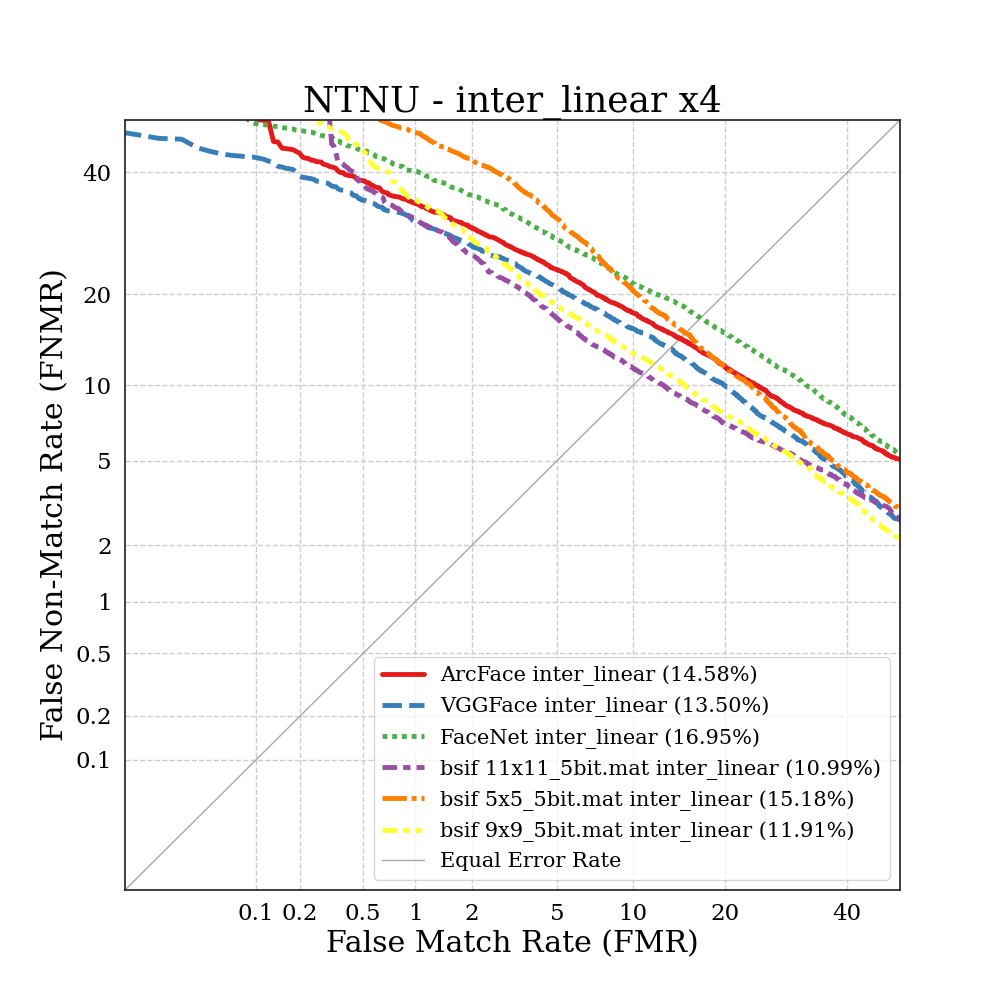}
\caption{DETs for MObBIO and NTNU dataset including periocular recognition systems based on traditional Inter-Linear resizing. The EER is showed in parenthesis for each method}. 
\label{DETs-mobio-ntu-linear}
\end{figure*}

Figs. \ref{DETs-mobio-ntu-area}, \ref{DETs-mobio-ntu-cubic}, and \ref{DETs-mobio-ntu-linear} show the DET curves results of periocular verification system with a standard resolution (No redimension) for MObBIO and NTNU dataset. 
A comparison with  Inter-area, cubic and lineal resized by x2, x3 and x4 is depicted. The results also show EERs for ArcFace, VGGFace, FaceNet and BSIF.

\end{document}